\documentclass[acmsmall]{acmart}
\usepackage{tabularx}
\usepackage{multirow}
\usepackage{subfigure}
\usepackage{color}
\usepackage{hyperref}
\hypersetup{hidelinks,
	colorlinks=true,
	allcolors=black,
	pdfstartview=Fit,
	breaklinks=true}
	
\AtBeginDocument{%
  \providecommand\BibTeX{{%
    \normalfont B\kern-0.5em{\scshape i\kern-0.25em b}\kern-0.8em\TeX}}}

\setcopyright{acmcopyright}
\copyrightyear{2021}
\acmYear{2021}
\acmDOI{}

\acmJournal{CSUR}
\acmVolume{0}
\acmNumber{0}
\acmArticle{0}
\acmMonth{0}

\begin{document}
\bibliographystyle{unsrt}
\title{Deep Learning-based Face Super-Resolution: A Survey}

\author{Junjun Jiang}

\affiliation{%
  \institution{School of Computer Science and Technology, Harbin Institute of Technology}
  \city{Harbin}
  \country{China}}
\email{jiangjunjun@hit.edu.cn}
\author{Chenyang Wang}
\affiliation{%
  \institution{School of Computer Science and Technology, Harbin Institute of Technology}
  \city{Harbin}
  \country{China}}
\email{wangchy02@hit.edu.cn}

\author{Xianming Liu}
\affiliation{%
  \institution{School of Computer Science and Technology, Harbin Institute of Technology}
  \city{Harbin}
  \country{China}
}
\email{csxm@hit.edu.cn}
\author{Jiayi Ma}
\affiliation{%
 \institution{Electronic Information School, Wuhan University}
 \city{Wuhan}
 \country{China}}

\renewcommand{\shortauthors}{J. Jiang and C. Wang, et al.}

\begin{abstract}
Face super-resolution (FSR), also known as face hallucination, which is aimed at enhancing the resolution of low-resolution (LR) face images to generate high-resolution (HR) face images, is a domain-specific image super-resolution problem. Recently, FSR has received considerable attention and witnessed dazzling advances with the development of deep learning techniques. To date, few summaries of the studies on the deep learning-based FSR are available. In this survey, we present a comprehensive review of deep learning-based FSR methods in a systematic manner\footnote{A curated list of papers and resources to face super-resolution at \url{https://github.com/junjun-jiang/Face-Hallucination-Benchmark}.}. First, we summarize the problem formulation of FSR and introduce popular assessment metrics and loss functions. Second, we elaborate on the facial characteristics and popular datasets used in FSR. Third, we roughly categorize existing methods according to the utilization of facial characteristics. In each category, we start with a general description of design principles, then present an overview of representative approaches, and then discuss the pros and cons among them. Fourth, we evaluate the performance of some state-of-the-art methods. Fifth, joint FSR and other tasks, and FSR-related applications are roughly introduced. Finally, we envision the prospects of further technological advancement in this field.

\end{abstract}



\maketitle

The research is supported by the National Natural Science Foundation of China (61971165, 61922027, 61773295), and the Fundamental Research Funds for the Central Universities.

\section{Introduction}
\label{intro}

Face super-resolution (FSR), a domain-specific image super-resolution problem, refers to the technique of recovering high-resolution (HR) face images from low-resolution (LR) face images. It can increase the resolution of an LR face image of low quality and recover the details. In many real-world scenarios, limited by physical imaging systems and imaging conditions, the face images are always low quality. Thus, with a wide range of applications and notable advantages, FSR has always been a hot topic since its birth in image processing and computer vision.

\begin{table*}
\caption{Summary of face super-resolution surveys since 2010.}

      \begin{tabularx}{\linewidth}{m{0.2cm}<{\centering} m{10.5cm}<{\centering} m{0.5cm}<{\centering} m{1.3cm}<{\centering}}
    \toprule
         \multirow{2}{*}{No.}&  \multirow{2}{*}{Survey title}  &  \multirow{2}{*}{Year} & \multirow{2}{*}{Venue} \\
         &  &    &  \\
         \midrule

         1 & A survey of face hallucination~\cite{FSRsurvey12012} & 2012&CCBR \\
         2 & A comprehensive survey to face hallucination~\cite{FSRsurvey2014}&2014&IJCV\\
         3 & A review of various approaches to face hallucination~\cite{survey2015} & 2015 &ICACTA \\
         4& Face super resolution: a survey ~\cite{FSRsurvey2017} & 2017 & IJIGSP \\
         5 & Super-resolution for biometrics: a comprehensive survey~\cite{survey2018super}  &2018 & PR\\
         6 &Face hallucination techniques: a survey~\cite{FSRsurvey2018} &2018 &CICT \\
         7 & Survey on GAN-based face hallucination with its model development~\cite{FSRsurvey2019} & 2019 &IET \\ 
         \bottomrule
    \end{tabularx}

    \label{tab:survey}
\end{table*}

The concept of FSR was first proposed in 2000 by Baker and Kanade~\cite{baker}, who are the pioneers of the FSR technique. They develop a multi-level learning and prediction model based on the Gaussian image pyramid to improve the resolution of an LR face image. Liu \emph{et al}.~\cite{liu2001two} propose to integrate a global parametric principal component analysis (PCA) model with a local nonparametric Markov random field (MRF) model for FSR. Since then, a number of innovative methods have been proposed, and FSR has become the subject of active research efforts. Researchers super-resolve the LR face images by means of global face statistical models~\cite{Gunturk2003,Wang2005Eig,Chakrabarti2007,Park2008,innerhofer2013convex,Liang2013,yang2013structured}, local patch-based representation methods~\cite{chang2004super,ma2010hallucinating,jung2011position,jiang2014noise,farrugia2017face,jiang2018context, shi2018hallucinating,chen2019robust,shi2019face, liu2020hallucinating,chen2021multi}, or hybrid ones~\cite{zhuang2007hallucinating,huang2010super}. These methods have achieved good performance, however, have trouble when meeting requirements in practice. With the rapid development of deep learning technique, attractive advantages over previous attempts have been obtained and have been applied into image or video super-resolution. Many comprehensive surveys have reviewed recent achievements in these fields, \emph{i.e.}, general image super-resolution surveys~\cite{SR1,SR2,SR3}, and video super-resolution survey~\cite{video2020}. Towards FSR, a domain-specific image super-resolution, a few surveys are listed in Table~\ref{tab:survey}. In the early stage of research, ~\cite{FSRsurvey12012,FSRsurvey2014,survey2015,FSRsurvey2017,survey2018super,FSRsurvey2018} provide a comprehensive review of traditional FSR methods (mainly including patch-based super-resolution, PCA-based methods, \emph{etc}.), while Liu \emph{et al}.~\cite{FSRsurvey2019} offer a generative adversarial network (GAN) based FSR survey. However, so far no literature review is available on deep learning super-resolution specifically for human faces. In this paper, we present a comparative study of different deep learning-based FSR methods.
\\
The main contributions of this survey are as follows:
\begin{itemize}
    \item The survey provides a comprehensive review of recent techniques for FSR, including problem definition, commonly used evaluation metrics and loss functions, the characteristics of FSR, benchmark datasets, deep learning-based FSR methods, performance comparison of state-of-the-arts, methods that jointly perform FSR and other tasks, and FSR-related applications.
\item The survey summarizes how existing deep learning-based FSR methods explore the potential of network architecture and take advantage of the characteristics of face images, as well as compare the similarities and differences among these methods.
\item The survey discusses the challenges and envisions the prospects of future research in the FSR field.
\end{itemize}

\begin{figure*}[t]
\includegraphics[width=\linewidth]{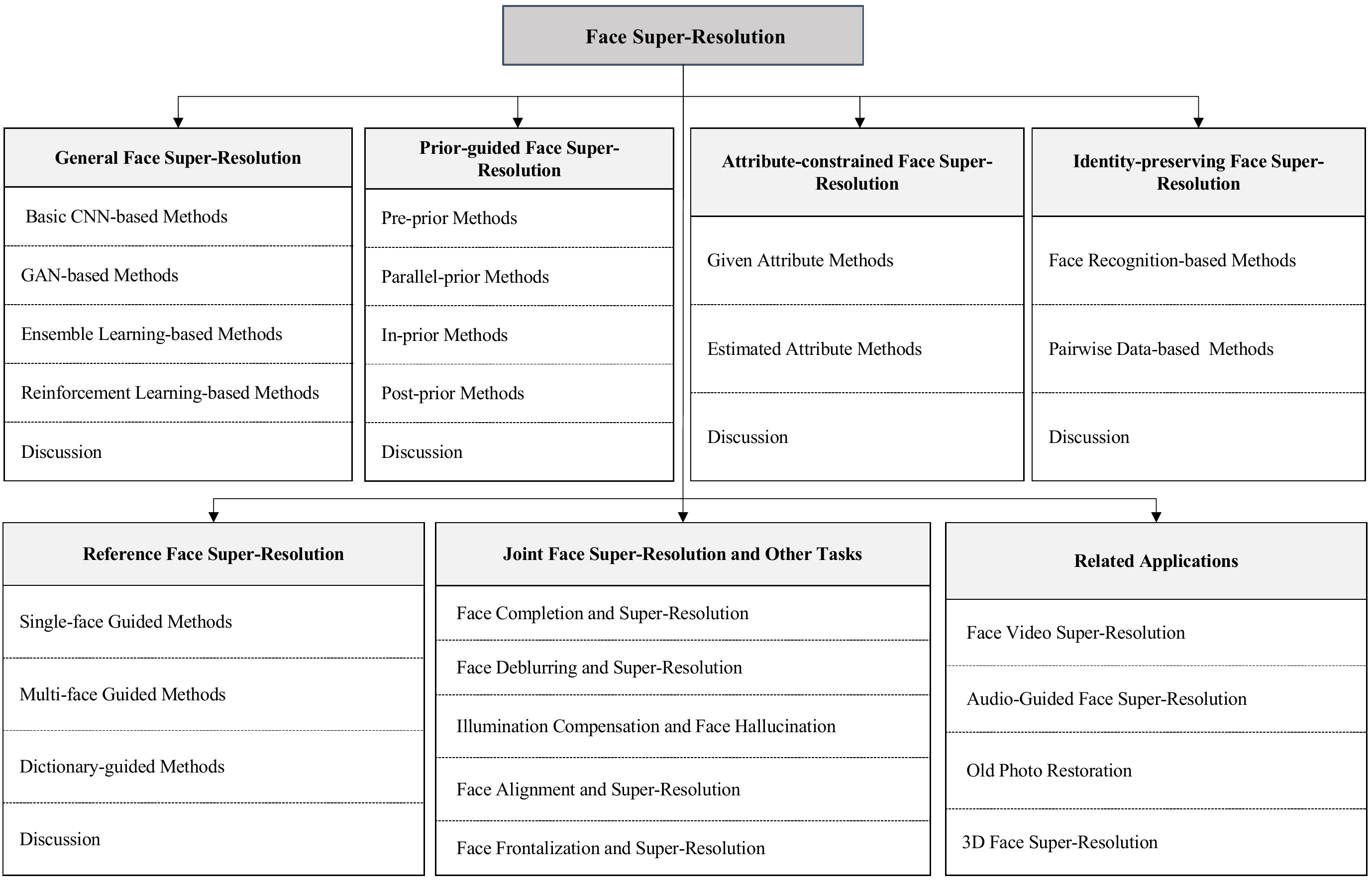}
\caption{The taxonomy of face super-resolution.}
\label{overall1}
\end{figure*}

In the following, we will cover the existing deep learning-based FSR methods and Fig.~\ref{overall1} shows the taxonomy of FSR. Section~\ref{Sec2} introduces the problem definition of FSR, and commonly used assessment metrics and loss functions. Section~\ref{Sec3} presents the facial characteristics (\emph{i.e.}, prior information, attribute information, and identity information) and reviews some mainstream face datasets. In Section~\ref{Sec4}, we discuss FSR methods. To avoid exhaustive enumeration and take facial characteristics into consideration, FSR methods are categorized according to facial characteristics used. In Section~\ref{Sec4}, five major categories are presented: general FSR methods, prior-guided FSR methods, attribute-constrained FSR methods, identity-preserving FSR methods, and reference FSR methods. Depending on the network architecture or the utilization of facial characteristics, every category is further divided into several subcategories. Moreover, Section \ref{Sec4} compares the performance of some state-of-the-art methods. Besides, Section~\ref{Sec4} also reviews some methods dealing with joint tasks and FSR-related applications. Section~\ref{Sec5} concludes the FSR and further discusses the limitations as well as envisions the prospects of further technological advancement.

\section{Background}\label{Sec2}
\subsection{Problem Definition}

FSR focuses on recovering the corresponding HR face image from an observed LR face image. The image degradation model $\Phi$ can be mathematically written as:
\begin{equation}  \label{eq1}
I_{\text{LR}} = \Phi(I_{\text{HR}}, \theta),
\end{equation}
where $\theta$ represents the model parameters including blurring kernel, downsampling operation, and noise, $I_{\text{LR}}$ is the observed LR face image, and $I_{\text{HR}}$ is the original HR face image. FSR is devoted to simulating the inverse process of the degradation model and recovers the $I_{\text{SR}}$ from $I_{\text{LR}}$, which can be expressed as:
\begin{equation}\label{eq2}
I_{\text{SR}}=\Phi ^{\text{-1}}(I_{\text{LR}}, \delta)=F(I_{\text{LR}}, \delta),
\end{equation}
where $F$ is the super-resolution model (inverse degradation model), $\delta$ represents the parameters of $F$, and $I_{\text{SR}}$ represents the super-resolved result. The optimization of $\delta$ can be defined as:
\begin{equation}\label{eq9}
\hat{\delta}=\underset{\delta}{\text{argmin}}\,\,\mathcal{L}(I_{\text{SR}}, I_{\text{HR}}),
\end{equation}
where $\mathcal{L}$ represents the loss between $I_{\text{SR}}$ and $I_{\text{HR}}$ and $\hat{\delta}$ is the optimal parameter of the trained model. In FSR, MSE loss and $\mathcal{L}_1$ loss are the most popular loss functions, and some models tend to use a combination of multiple loss functions, which will be reviewed in Section~\ref{Loss}.

The degradation model and parameters are all unavailable in a real-world environment, and $I_{\text{LR}}$ is the only given information. To simulate the image degradation process, researchers tend to use mathematical models to generate some LR and HR pairs to train the model. The simplest mathematical model is
\begin{equation} \label{eq3}
I_{\text{LR}}=(I_{\text{HR}})\downarrow_{s},
\end{equation}
where $\downarrow$ denotes the downsampling operation, and $s$ is the scaling factor. However, this pattern is too simple to match the real-world degradation process. To better mimic the real degradation process, researchers design a degradation process with the combination of many operations (\emph{e.g.}, downsampling, blur, noise, and compression) as follows:
\begin{equation}\label{eq5}
I_{\text{LR}}=J((I_{\text{HR}}\otimes k)\downarrow_{s} + \,n),
\end{equation}
where $k$ is the blurring kernel, $\otimes$ represents the convolutional operation, $n$ denotes the noise, and $J$ denotes the image compression. Various combinations of different operations are used in FSR. They include the widely used bicubic model~\cite{URDGN,FSRNet,DIC}, as well as the general degradation model used for blind FSR \cite{ASFFNet,DFDNet,GFRNet}. However, they are not introduced in detail in this survey.

\subsection{Assessment Metrics and Loss Functions}\label{Loss}

In deep learning-based FSR methods, the loss function which measures the difference between $I_{\text{HR}}$ and $I_{\text{SR}}$ plays an important role in guiding the network training. Upon acquiring the trained network, the reconstruction performance of these methods can be evaluated by the assessment metrics. The preferences of different loss functions are different. For example, $\mathcal{L}_2$ loss tends to produce the result that is faithful to the original image (high PSNR value), and the perceptual and adversarial losses will generate subjectively pleasing results (low FID~\cite{FID} and LPIPS~\cite{lpips} values). In practice, we can choose the appropriate loss function according to the needs. Considering the relationship between loss functions and assessment metrics, we introduce them together in this section.
\subsubsection{Image Quality Assessment}
Generally, two main methods of quality evaluation are subjective and objective evaluation. Subjective evaluation relies on the judgement of humans, and tends to invite readers or interviewers to see and assess the quality of the generated images, leading to results always consistent with human perception but time-assuming, inconvenient and expensive. In contrast, the objective evaluation mainly utilizes statistical data to reflect the quality of the generated images. In general, the objective evaluation methods usually produce different results from subjective evaluation metrics, because the starting point of objective evaluation methods is mathematics instead of human visual perception, which leaves the assessment image quality in dispute. Here, we introduce some popular assessment metrics.

\textbf{Peak Signal-to-Noise Ratio (PSNR):} PSNR is a commonly used objective assessment metric in FSR. Given $I_{\text{HR}}$ and $I_{\text{SR}}$, the mean square error (MSE) between them is firstly calculated, then the PSNR is obtained,
\begin{equation}
\text{MSE} = \frac{\text{1}}{hwc}\left\|I_{\text{SR}}-I_{\text{HR}}\right\|^{2}_{2},
\end{equation}
\begin{equation}
\text{PSNR} = 10 \, log_{\text{10}}(\frac{\text{M}^{2}}{\text{MSE}}),
\end{equation}
where $h$, $w$, and $c$ denote the height, width, and channel of the image, and $\text{M}$ is the maximum possible pixel value (\emph{i.e.}, 255 for 8-bit images). The smaller the pixel-wise difference of the two images, the higher the PSNR. In this pattern, PSNR focuses on the distance between every pair of pixels in two images, which is inconsistent with human perception, resulting in poor performance when human perception is more important.

\textbf{Structural Similarity Index (SSIM):} SSIM~\cite{SSIM} is also a popular objective assessment metric that measures the structural similarity between two images. To be specific, SSIM measures similarity from three aspects: luminance, contrast, and structure. Given $I_{\text{HR}}$ and $I_{\text{SR}}$, SSIM is obtained by
\begin{equation}
\text{SSIM}=l(I_{\text{HR}}, I_{\text{SR}})*C(I_{\text{HR}}, I_{\text{SR}})*S(I_{\text{HR}}, I_{\text{SR}}),
\end{equation}
where $l(I_{\text{HR}}, I_{\text{SR}})$, $C(I_{\text{HR}}, I_{\text{SR}})$ and $S(I_{\text{HR}}, I_{\text{SR}})$ denote the similarity of the
luminance, contrast and structure. SSIM varies from 0 to 1. The higher the structural similarity of the two images, the larger the SSIM. Considering the uneven distribution of the image, SSIM is not reliable enough. Thus, multi-scale structural similarity index measure (MS-SSIM)~\cite{MS-SSIM} is proposed, which divides the image into multiple windows, first assesses SSIM for every window separately, and then converges them to obtain MS-SSIM.

\textbf{Learned Perceptual Image Patch Similarity (LPIPS):} LPIPS~\cite{lpips} measures the distance between two images in a deep feature space. LPIPS is more in line with human judgement than PSNR and SSIM. The more similar the two images, the smaller the LPIPS.

\textbf{Fr\'echet Inception Distance (FID):} In contrast to PSNR and SSIM, FID~\cite{FID} focuses on the difference between $I_{\text{HR}}$ and $I_{\text{SR}}$ in a distribution-wise manner, and it is always applied to assess the visual quality of face images. The better the visual quality, the smaller the FID.

\textbf{Natural Image Quality Evaluator (NIQE):} NIQE~\cite{NIQE} is a no-reference metric that measures the distance between two multivariate Gaussian models fitting natural images and the evaluated images without ground truth images. Specifically, the fitting of multivariate Gaussian model is based on the quality-aware features derived from the natural scene statistic model. The better the visual quality, the smaller the NIQE.

\textbf{Mean Opinion Score (MOS):} MOS is a commonly used subjective assessment metric, in contrast to the above objective quantitative metrics. To obtain the MOS, human raters are asked to assign perceptual quality scores to the tested images. Finally, MOS is obtained by calculating the arithmetic mean ratings assigned by human raters. When the number of human raters is small, MOS would be biased while MOS would be faithful enough when the number of human raters is large.

\subsubsection{Loss Functions} Initially, pixel-wise $\mathcal{L}_{2}$ loss (also known as MSE loss) is popular, however, researchers then find that models based on $\mathcal{L}_{2}$ loss tend to generate smooth results. Then many kinds of loss functions are employed, such as pixel-wise $\mathcal{L}_{1}$ loss, SSIM loss, perceptual loss, adversarial loss, \emph{etc}.

\textbf{Pixel-wise Loss: }Pixel-wise loss measures the distance between the two images at pixel level, including $\mathcal{L}_{1}$ loss that calculates the mean absolute error, $\mathcal{L}_{2}$ loss that calculates the mean square error, Huber loss~\cite{CAGFace} and Carbonnier penalty function~\cite{Car}. With the constrain of the pixel-wise loss, the obtained $I_{\text{SR}}$ can be close enough to the $I_{\text{HR}}$ on the pixel value. From the definition, $\mathcal{L}_{2}$ loss is sensitive to large errors but indifferent to small errors, while $\mathcal{L}_{1}$ loss treats them equally. Therefore, $\mathcal{L}_{1}$ loss has advantages in improving the performance and convergence over $\mathcal{L}_{2}$ loss. Overall, pixel-wise loss can force the model to improve PSNR, but the generated images are always over-smooth and lack high-frequency details.

\textbf{SSIM Loss: }Similar to pixel-wise loss, SSIM loss is designed to improve the structure similarity between super-resolved image and the original HR one:
\begin{equation}
\mathcal{L}_{\text{SSIM}}(I_{\text{HR}}, I_{\text{SR}})=\frac{1}{2}\left(1-F_{\text{SSIM}}(I_{\text{HR}}, I_{\text{SR}})\right),
\end{equation}
where $F_{\text{SSIM}}$ denotes the function of SSIM. Except for SSIM loss, multi-scale SSIM loss can calculate SSIM loss at different scales.

\textbf{Perceptual Loss:} To improve the perceptual quality, one solution is to minimize the perceptual loss:
\begin{equation}
\mathcal{L}_{\text{Perceptual}}(I_{\text{HR}},I_{\text{SR}},\Psi,l)=
\left\|\Psi^{l}(I_{\text{HR}})-\Psi^{l}(I_{\text{SR}})\right\|_{2},
\end{equation}
where $\Psi$ is the pretrained network and $l$ is the $l$-th layer. In essence, the perceptual loss measures the distance between the features extracted from $\Psi$ (\emph{e.g.}, VGG~\cite{VGG}), and it can evaluate the difference at the semantic level. Perceptual loss encourages the network to generate $I_{\text{SR}}$ that is more perceptually similar to $I_{\text{HR}}$. The $I_{\text{SR}}$ predicted by the model with perceptual loss always looks more pleasant but usually has lower PSNR than those pixel-wise loss-based methods.

\textbf{Adversarial Loss: }Adversarial loss, proposed in generative adversarial network (GAN)~\cite{GAN}, is also widely used in FSR. For details, GAN is comprised of two models: a generator (G) and a discriminator (D). In FSR, GAN can be described as follows: G is the super-resolution model which generates the super-resolved face with an LR face image as input, and D discriminates whether the output result is generated or real. In the training phase, G and D are trained alternatively. Early methods ~\cite{URDGN,TDN} use cross entropy-based adversarial loss expressed as follows:
\begin{equation}
\mathcal{L}_{\text{G}}(I_{\text{SR}})=-\text{log}(\mathcal{D}(I_{\text{SR}})),
\end{equation}
\begin{equation}
\mathcal{L}_{\text{D}}(I_{\text{HR}},I_{\text{SR}})=-\text{log}(\mathcal{D}(I_{\text{HR}}))-\text{log}(1-\mathcal{D}(I_{\text{SR}})),
\end{equation}
where $\mathcal{L}_{\text{D}}$ and $\mathcal{L}_{\text{G}}$ denotes the loss function of D and G, respectively, $\mathcal{D}$ denotes the function of D, and $I_{\text{HR}}$ is randomly sampled from HR training samples. However, the model trained with this adversarial loss is always unstable and may cause model collapse. Therefore, Wasserstein GAN~\cite{WGAN} and WGAN-GP~\cite{WGAN-GP} are proposed to alleviate the training difficulties. The model trained with adversarial loss tends to introduce artificial details, leading to worse PSNR and SSIM but pleasing visual quality with smaller FID.

\textbf{Cycle Consistency Loss:} Cycle consistency loss is proposed by CycleGAN~\cite{CycleGAN}. In CycleGAN-based FSR, two cooperated models are used: a super-resolution model super-resolves the $I_{\text{LR}}$ to recover the $I_{\text{SR}}$, and a degradation model downsamples the $I_{\text{SR}}$ back to $I_{\text{LR}^{'}}$. In turn, the degradation model downsamples the HR face image to obtain $I_{\text{HLR}}$, and then the super-resolution model recovers the $I_{\text{HLR}}$ to generate $I_{\text{HR}^{'}}$. The cycle consistent loss is aimed to keep the consistency between $I_{\text{LR}}$ ($I_{\text{LR}^{'}}$) and $I_{\text{HR}}$ ($I_{\text{HR}^{'}}$),
\begin{equation}
\mathcal{L}_{\text{Cycle}}(I_{\text{LR}}, I_{\text{LR}^{'}}, I_{\text{HR}}, I_{\text{HR}^{'}})=\left\| I_{\text{LR}}-I_{\text{LR}^{'}}\right\|_2+ \left\| I_{\text{HR}}-I_{\text{HR}^{'}}\right\|_2.
\end{equation}

In addition to the above loss functions, many other loss functions are also used in FSR, including style loss~\cite{SL}, feature match loss~\cite{MSFM}, \emph{etc}. Due to the limitation of space, we do not introduce them in detail.

\section{Characteristics of Face Images}\label{Sec3}

\begin{figure*}[t]
    \centering
    \includegraphics[width=\linewidth]{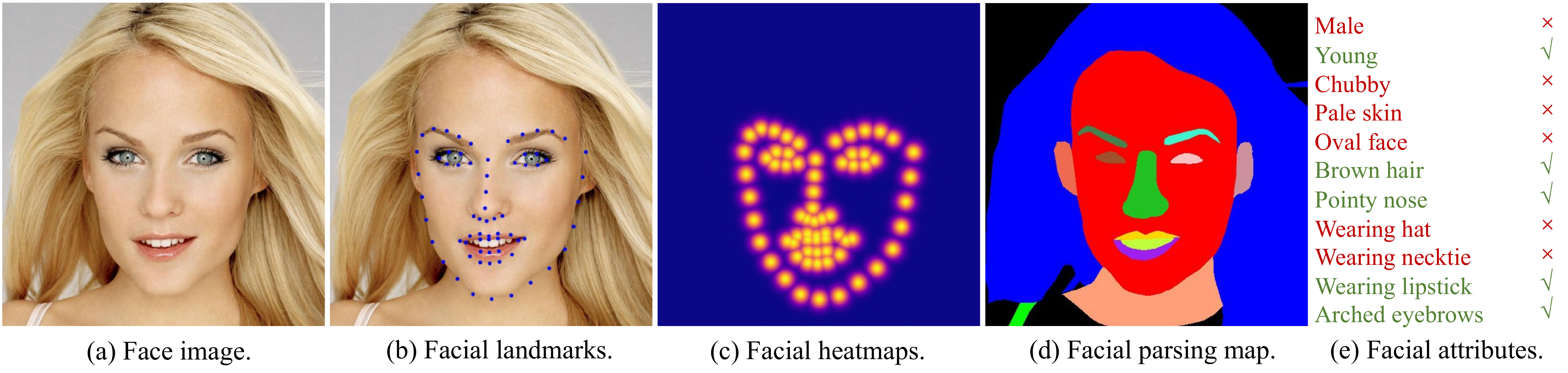}
    \caption{Facial characteristics.}
    \label{fig:prior}
\end{figure*}
Human face is a highly structured object with its own unique characteristics, which can be explored and utilized in FSR task. In this section, we simply introduce these facial characteristics.
\subsection{Prior Information}
As shown in Fig. \ref{fig:prior}, structural priors can be found in face images, such as facial landmarks, facial heatmaps and facial paring maps.
\begin{itemize}
\item \emph{Facial landmarks}: These locate the key points of facial components. The number of landmarks varies in different datasets, such as CelebA~\cite{celeba}, which provides five landmarks while Helen~\cite{helen} offers 194 landmarks.
\item \emph{Facial heatmaps}: These are generated from facial landmarks. Facial landmarks give accurate points of the facial components, while heatmaps give the probability of the point being a facial landmark. To generate the heatmaps, every landmark is represented by a Gaussian kernel centered on the location of the landmark.
\item \emph{Facial parsing maps}: These are semantic segmentation maps of face images separating the facial components from face images, including eyes, nose, mouth, skin, ears, hair, and others.

\end{itemize}
These face structure prior information can provide the location of facial components and facial structure information. We can expect to recover more reasonable target face images if we incorporate these prior knowledge to regularize or guide the FSR models.

\begin{table*}
\caption{Summary of public face image datasets for FSR.}
\centering
\begin{tabularx}{\linewidth}{m{4cm}<{\centering}X<{\centering}X<{\centering}X<{\centering}X<{\centering}X<{\centering}}
\toprule
\multirow{2}{*}{Dataset}  & \multirow{2}{*}{Number}  & \multirow{2}{*}{\#Attributes} & \multirow{2}{*}{\#Landmarks} & \multirow{2}{*}{Parsing maps} & \multirow{2}{*}{Identity}\\
& & & & &\\
\midrule
CelebA~\cite{celeba}& 202,599&40&5& $\times$  &\checkmark\\
CelebAMask-HQ~\cite{CelebAMask-HQ}&30,000&$\times$&$\times$& \checkmark&  $\times$ \\
Helen~\cite{helen} & 2,330 &  $\times$ &194& \checkmark& $\times$ \\
FFHQ~\cite{StyleGAN}& 70,000  &  $\times$ &68 & $\times$  & $\times$\\
AFLW~\cite{AFLW} & 25,993   & $\times$   & 21&$\times$  &$\times$\\
300W~\cite{300W1} & 3,837    &   $\times$ &68 & $\times$ & $\times$\\
LS3D-W~\cite{LS3D-W} & 230,000     & $\times$ &68 & $\times$ & $\times$\\
Menpo~\cite{Menpo} & 9,000       & $\times$ &68 & $\times$ & $\times$\\
LFW~\cite{LFW} & 13,233    &73 &$\times$ & $\times$ & \checkmark\\
LFWA~\cite{LFW-a} & 13,233    &    40 &$\times$ & $\times$ & \checkmark\\
VGGFace~\cite{VGGFaces} & 3,310,000      & $\times$ &$\times$ & $\times$ & \checkmark\\
\bottomrule
\end{tabularx}

\label{tab:dataset}
\end{table*}

\subsection{Attribute Information}
Second, the attributes, such as gender, hair color, and others, are the affiliated features of face images and can be seen as semantic-level information. In FSR, because of one-to-many maps from LR images to HR ones, the recovered face image may contain artifacts and even wrong attributes. For example, the face in the recovered result does not wear but the ground truth wears eyeglasses. At this time, attribute information can remind the network which attribute should be covered in the result. From a different perspective, attribute information also contains facial details. Taking eyeglasses as an example, the attribute of wearing eyeglasses provides the details of the facial eyes. We provide a concise example of attribute information in Fig.~\ref{fig:prior}. Moreover, these attributes are always binary in the face dataset, 1 denotes that the face image has the attribute, while 0 means there is no such information.
\subsection{Identity Information}
Third, every face image corresponds to a person, which is enabled by identity information. This type of information is always used for keeping the identity consistency between the super-resolved result and the ground truth. On the one hand, the person should not be changed after super-resolution visually. On the other hand, FSR should facilitate the performance of face recognition. Similar to attribute information, identity also offers high-level constraints to the FSR task and is beneficial to face restoration.

\subsection{Datasets for FSR}
In recent years, many face image datasets are used for FSR, which differ in many aspects, \emph{e.g.}, the number of samples, facial characteristics contained and others. In Table~\ref{tab:dataset}, we list a number of commonly used face image datasets and simply indicate their amount and the facial characteristics offered. For parsing maps and identity, we only present whether they are provided or not, while for attributes and landmarks, we offer the specific amount. Aside from these datasets, many other face datasets are used in FSR, including CACD200~\cite{CACD2000}, VGGFace2~\cite{VGGFace2}, UMDFaces~\cite{UMDFaces}, CASIA-WebFace~\cite{CASIA-WebFace}, and others. It is worth noting that all above-mentioned datasets only provide HR face images. If we want to use them for training and evaluating any super-resolution model, we need to generate the corresponding LR face images using the degradation model introduced in Section ~\ref{Sec2}.

\section{FSR Methods}\label{Sec4}

At present, various deep learning FSR methods have been proposed. On the one hand, these methods tap the potential of the efficient network for FSR regardless of facial characteristics, \emph{i.e.}, developing a basic convolution neural network (CNN) or generative adversarial network (GAN) for face reconstruction. On the other hand, some approaches focus on the utilization of facial characteristics, \emph{e.g.}, using structure prior information to facilitate face restoration and so on. Furthermore, some recently proposed models introduce additional high-quality reference face images to assist the restoration. Here, according to the type of face image special information used, we divide FSR methods into five categories: general FSR, prior-guided FSR, attribute-constrained FSR, identity-preserving FSR, and reference FSR. In this section, we concentrate on every kind of FSR method and introduce each category in detail.

\subsection{General FSR}
\begin{figure*}[t]
    \centering
    \includegraphics[width=0.95\linewidth]{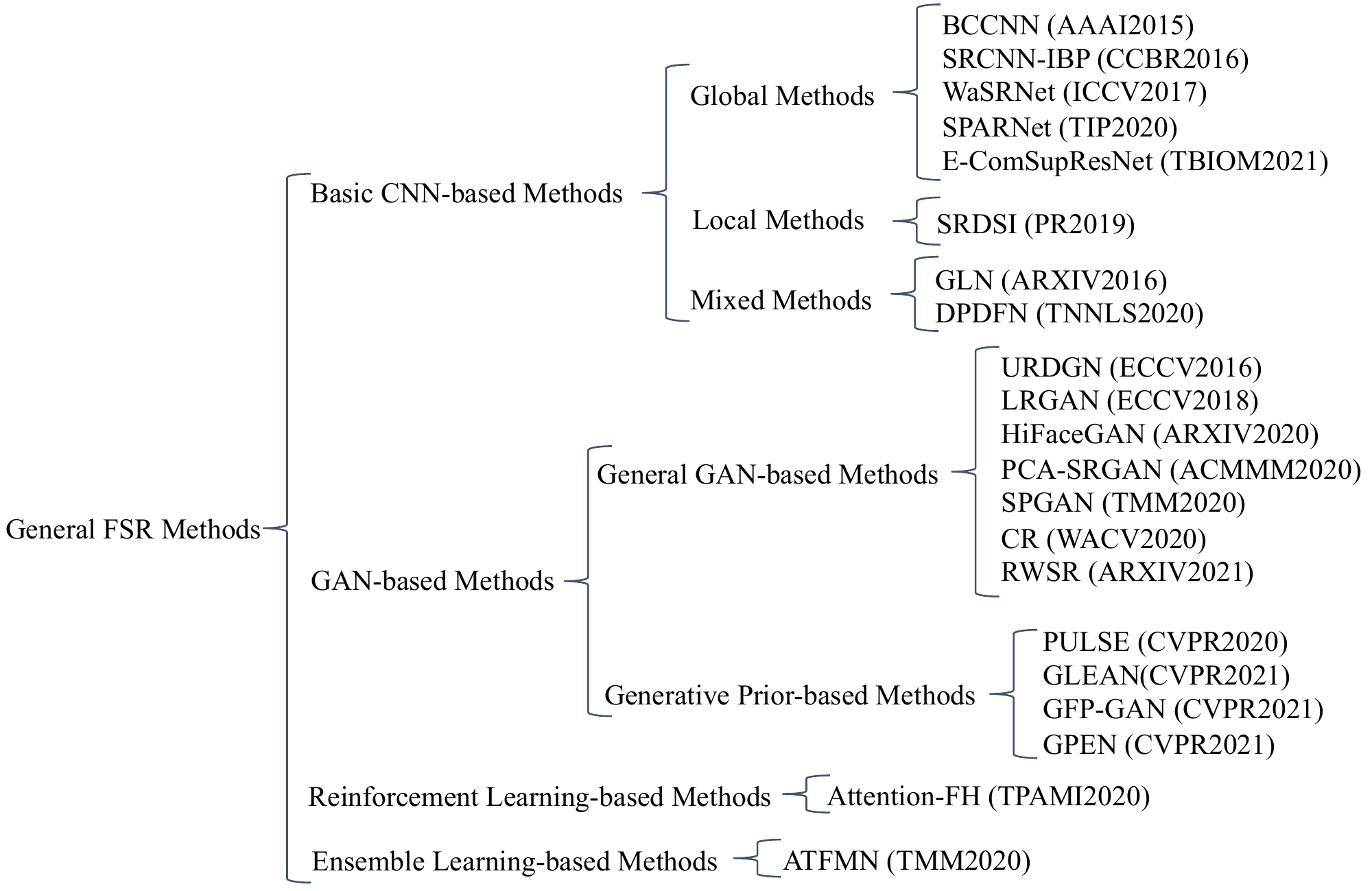}
    \caption{Overview of general FSR methods.}
    \label{general1}

\end{figure*}

General FSR methods mainly focus on designing an efficient network and exploit the potential of efficient network structure for FSR without any facial characteristics. In the early days, most of these methods are based on CNN and incorporate various advanced architectures (including back projection, residual network, spatial or channel attention, \emph{etc}.), to improve the representation ability of the network. Since then, many FSR methods by using advanced networks have been proposed. We divide general FSR methods into four categories: basic CNN-based methods, GAN-based methods, reinforcement learning-based methods, and ensemble learning-based methods. Aiming to present a clear and concise overview, we summarize the general FSR methods in Fig.~\ref{general1}.

\subsubsection{Basic CNN-based Methods}
Inspired by the pioneer deep learning general image super-resolution method \cite{SRCNN}, some researchers also propose to incorporate the CNN network into the FSR task. Depending on whether they consider the global information and local differences, we can further divide the basic CNN-based methods into three categories: global methods that feed the entire face into the network and recover face images globally, local methods that divide face images into different components and then recover them, and mixed methods that recover face images locally and globally.

\textbf{Global Methods:} In the early years, researchers treat a face image as a whole and recover it globally. Inspired by the strong representative ability of CNN, bi-channel convolutional neural network (BCCNN)~\cite{BCCNN} and ~\cite{SRCNNFace} directly learn a mapping from LR face images to HR ones. Then, benefiting from the performance gain of iterative back projection (IBP) in general image super-resolution, Huang \emph{et al}.~\cite{SRCNN-IBP} introduce IBP to FSR as an extra post-processing step, developing the SRCNN-IBP method. After that, the thought of back projection is generally used in FSR~\cite{RBPNet1,SCPN}. Later on, channel and spatial attention mechanisms greatly improve the general image super-resolution methods, which inspires researchers to explore their utilization in FSR. Thus, a number of innovative methods integrating the attention mechanism are proposed~\cite{9248111,E-ComSupResNet,SPARNet}. In these works, two representative methods are E-ComSupResNet~\cite{E-ComSupResNet} that introduces a channel attention mechanism and SPARNet~\cite{SPARNet} which has a well-designed spatial attention for FSR. Besides that, many researchers design the cascaded model and exploit multi-scale information to improve the restoration performance~\cite{CDFH,FHCNN,SGEN}.

It is observed that super-resolution in the image domain produces smooth results without high-frequency detail. Considering that wavelet transform can represent the textural and contextual information of the images, WaSRNet~\cite{WaveletSRNet} and \cite{2020Learning} transform face images into wavelet coefficients and super-resolve the face images in the wavelet coefficient domain to avoid over-smooth results.

\textbf{Local Methods:} Global methods can capture global information but cannot well recover the face details. Thus, local methods are developed to recover different parts of a face image differently. Super-resolution technique based on de?nition-scalable inference (SRDSI)~\cite{SRDSI} decomposes the face into a basic face with low-frequency and a compensation face with high-frequency through PCA. Then, SRDSI recovers the basic face and the compensation face with very deep convolutional network (VDSR)~\cite{VDSR} and sparse representation respectively. Finally, the two recovered faces are fused. After that, many patch-based methods have been proposed~\cite{PFH,FHDTN,RDCN}, all of which divide face images into several patches and train models for recovering the corresponding patches.

\textbf{Mixed Methods:} Considering that global methods can capture global structure but ignore local details while local methods focus on local details but lose global structure, a line of research naturally combines global and local methods for capturing global structure and recovering local details simultaneously. At first, global-local network \cite{GLN, lu2020global} develop a global upsampling network to model global constraints and a local enhancement network to learn face-specific details. To simultaneously capture global clues and recover local details, dual-path deep fusion network (DPDFN)~\cite{DPDFN} constructs two individual branches for learning global facial contours and local facial component details, and then fuses the result of the two branches to generate the final SR result.

\subsubsection{GAN-based Methods}\label{SPGAN}
Compared with CNN-based methods that utilize pixel-wise loss and generate smooth face images, GAN, first proposed by Goodfellow \emph{et al}.~\cite{GAN}, which can be applied to generate realistic-looking face images with more details, inspires researchers to design GAN-based methods. At first, researchers focus on designing various GANs to learn from paired or unpaired data. In recent years, how to utilize a pretrained generative model to boost FSR has attracted increasing attention. Therefore, GAN-based methods can be divided into general GAN-based methods and generative prior-based methods.

\textbf{General GAN-based Methods:}
In the early stage, Yu \emph{et al}.~\cite{URDGN} develop ultra-resolving face images by discriminative generative networks (URDGN), which consists of two subnetworks: a discriminative model to distinguish a real HR face image or an arti?cially super-resolved output, and a generative model to generate SR face images to fool the discriminative model and match the distribution of HR face images. MLGE~\cite{MLGE} not only designs discriminators to distinguish face images but also applies edge maps of the face images to reconstruct HR face images. Recently, HiFaceGAN~\cite{HiFaceGAN1} and the works of \cite{FFVFSR,IRN,WGANFSR,FCGAN} also super-resolve face images with generative models. Instead of directly feeding the whole face images into the discriminator, PCA-SRGAN~\cite{PCA-SRGAN} decomposes face images into components by PCA and progressively feeds increasing components of the face images into the discriminator to reduce the learning difficulty of the discriminator. The commonality of these types of GAN is that the discriminator outputs a single probability value to characterize whether the result is a real face image. However, Zhang \emph{et al}. \cite{SPGAN} assume that a single probability value is too fragile to represent a whole image, thus they design a supervised pixel-wise GAN (SPGAN) whose discriminator outputs a discriminative matrix with the same resolution as the input images, and design a supervised pixel-wise adversarial loss, thus recovering more photo-realistic face images.

The above methods rely on the artificial LR and HR pairs generated by a known degradation. However, the quality of the real-world LR image is affected by a wide range of factors such as the imaging conditions and the imaging system, leading to the complicated unknown degradation of real LR images. The gap between real LR images and artificial LR ones is large and will inevitably decrease the performance when applying methods trained on the artificial pairs to real LR images~\cite{2019facedata}. To settle this problem, real-world super-resolution (RWSR)~\cite{RWSR} first estimates the parameters from real LR faces, such as the blur kernel, noise, and compression, and then generates the LR and HR face image pairs with estimated parameters for the training of the model.

LRGAN~\cite{LRGAN} proposes to learn the degradation before super-resolution from unpaired data. It designs a high-to-low GAN to learn the real degradation process from unpaired LR and HR face images and create paired LR and HR face images for training low-to-high GAN. Specifically, with HR face images as input, the high-to-low GAN generates LR face images (GLR) that should belong to the real LR distribution and be close to the corresponding downsampled HR face images. Then, for low-to-high GAN, GLRs are fed into the generator to recover the SR results which have to be close to HR face images and match the real HR distribution. Goswami \emph{et al}.~\cite{RSR} further develop a robust FSR method and Zheng \emph{et al}. ~\cite{SDOT-CycleGAN} utilize semi-dual optimal transport to guide model learning and develop semi-dual optimal transport CycleGAN. Considering that discrepancies between GLRs in the training phase and real LR face images in the testing phase still exist, researchers introduce the concept of characteristic regularization (CR)~\cite{CR}. Different from LRGAN, CR transforms the real LR face images into artificial LR ones and then conducts super-resolution reconstruction in the artificial LR space. Based on CycleGAN, CR learns the mapping between real LR face images and artificial LR ones. Then, it uses the artificial LR face images generated from real LR ones to fine-tune the super-resolution model, which is pretrained by the artificial pairs.

\textbf{Generative prior-based methods:}
Recently, many face generation models, such as popular StyleGAN~\cite{StyleGAN}, StyleGAN v2 \cite{stylegan2}, ProGAN~\cite{progan}, StarGAN~\cite{stargan}, \emph{etc.}, have been proposed and they are capable of generating faithful faces with a high degree of variability. Thus, more and more researchers explore the generative prior of pretrained GAN.

The first generative prior-based FSR method is PULSE~\cite{Pulse}. It formulates FSR as a generation problem to generate high-quality SR face image so that the downsampled SR result is close to LR face image. Mathematically, the problem can be expressed as:
\begin{equation}
min_{G}\left\|G(z)\downarrow_{s}-I_{\text{LR}}\right\|,
\end{equation}
where $z$ is a randomly sampled latent vector and the input of the pretrained StyleGAN~\cite{StyleGAN}, $\downarrow$ is the downsampling operation, $s$ is the downsampling factor, and $G$ denotes the function of the generator. PULSE solves FSR from a new perspective and this inspires many other works.

However, the latent code $z$ in PULSE is randomly sampled and in low dimension, making the generated images lose important spatial information. To overcome this problem, GLEAN~\cite{GLEAN}, CFP-GAN~\cite{CFP-GAN} and GPEN~\cite{2021GAN} are developed. Rather than directly employing the pretrained StyleGAN~\cite{StyleGAN}, they develop their own networks and embed the pretrained generation network of StyleGAN~\cite{StyleGAN} into their own networks to incorporate the generative prior. To maintain faithful information, they not only obtain latent code by encoding LR face images instead of randomly sampling, but also extract multi-scale features from LR face images and fuse the features into the generation network. In this way, the generative prior provided by the pretrained StyleGAN can be fully utilized and the important spatial information can be well maintained.

\subsubsection{Reinforcement Learning-based Methods}
Deep learning-based FSR methods learn the mapping from LR face images to HR ones, but ignore the contextual dependencies among the facial parts. Cao \emph{et al}. propose to recurrently discover facial parts and enhance them by fully exploiting the global inter-dependency of the image, then attention-aware face hallucination via deep reinforcement learning (Attention-FH) is proposed~\cite{Attention-FH1}. Specifically, Attention-FH has two subnetworks: a policy network that locates the region that needs to be enhanced in the current step, and a local enhancement network that enhances the selected region.

\subsubsection{Ensemble Learning-based Methods}
CNN-based methods utilize pixel-wise loss to recover face images with higher PSNR and smoother details while GAN-based methods can generate face images with lower PSNR but more high-frequency details. To combine the advantages of different types of methods, ensemble learning is used in adaptive threshold-based multi-model fusion network (ATFMN)~\cite{ATFMN}. Specifically, ATFMN uses three models (CNN-based, GAN-based, and RNN-based) to generate candidate SR faces, and then fuses all candidate SR faces to reconstruct the final SR result. In contrast to previous approaches, ATFMN exploits the potential of ensemble learning for FSR instead of focusing on a single model.

\subsubsection{Discussion} Here we discuss the pros and cons among these sub-categories in general FSR methods. From a global perspective, the difference between CNN-based and GAN-based methods relies on adversarial learning. CNN-based methods tend to utilize pixel-wise loss, leading to higher PSNR and smoother results, while GAN-based methods might recover visually pleasing face images with more details but lower PSNR. Each of them has its own merits. Compared with them, ensemble learning-based method can combine their advantages to make up their deficiencies by integrating multiple models. However, ensemble learning inevitably results in the increase of memory, computation and parameters. Reinforcement learning-based methods recover the attentional local regions by sequentially searching, and consider the contextual dependency of patches from a global perspective, which brings improvement of performance but needs much more training time and computational cost.

\begin{figure*}[t]
\begin{center}

\subfigure[Framework of pre-prior methods.]{
    \begin{minipage}[t]{0.5\linewidth}
    \vspace{-0.5cm}
    \centering
        \includegraphics[width=\linewidth]{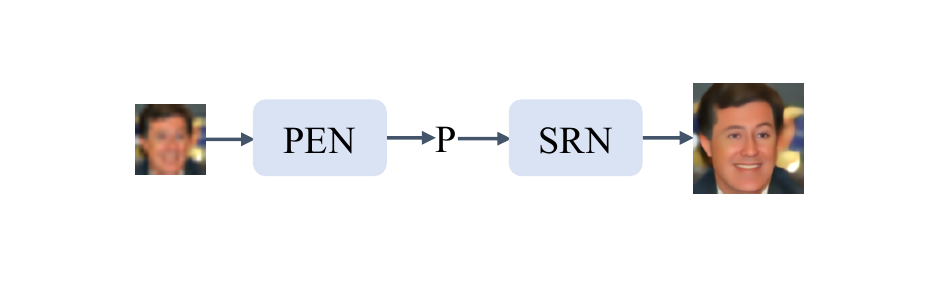}
    \end{minipage}%
}
\subfigure[Framework of parallel-prior methods.]{
    \begin{minipage}[t]{0.5\linewidth}
    \vspace{-0.5cm}
    \centering
        \includegraphics[width=\linewidth]{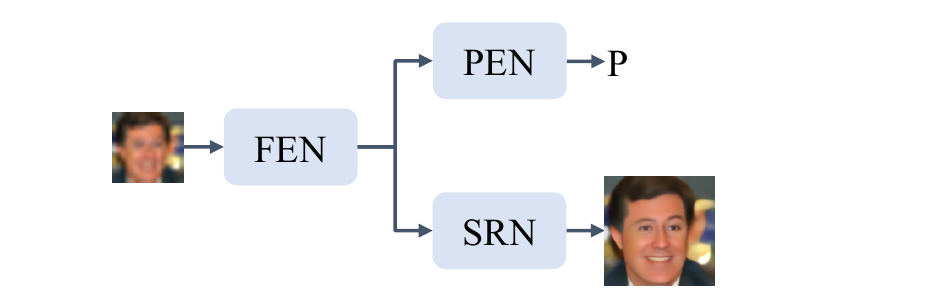}
    \end{minipage}%
}%

\quad
 \subfigure[Framework of in-prior methods.]{
   \begin{minipage}[t]{0.5\linewidth}
    \vspace{-0.5cm}
        \centering
        \includegraphics[width=\linewidth]{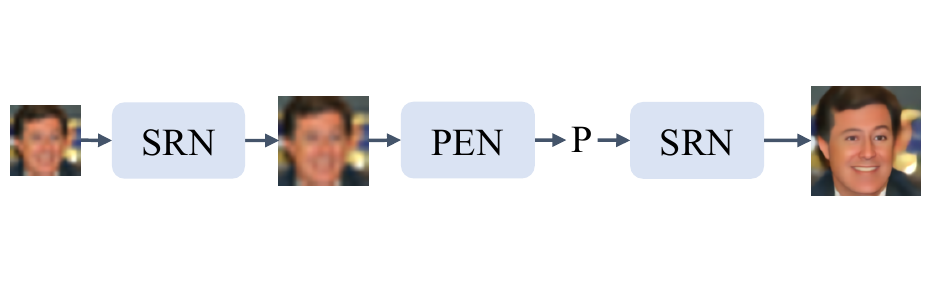}
    \end{minipage}%
}%
\subfigure[Framework of post-prior methods.]{
    \begin{minipage}[t]{0.5\linewidth}
    \vspace{-0.5cm}
    \centering
        \includegraphics[width=\linewidth]{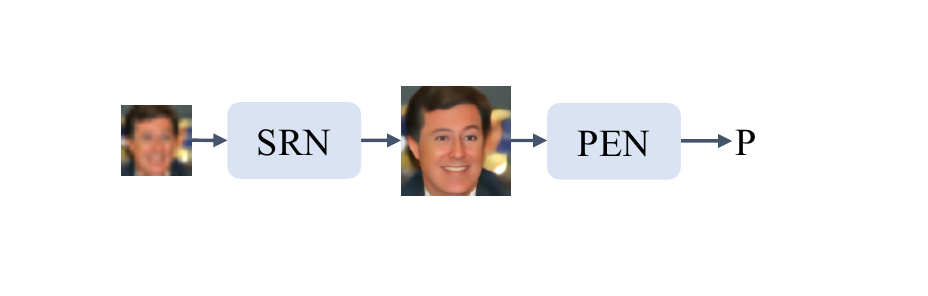}
    \end{minipage}%
}%

\quad
\end{center}
   \caption{Four frameworks of prior-guided FSR methods. PEN is prior estimation network, SRN is super-resolution network, FEN is a feature extraction network, and P is prior information.}
\label{prior1}
\end{figure*}
\subsection{Prior-guided FSR}\label{prior_guided_methods}
General FSR methods aim to design efficient networks. Nevertheless, as a highly structured object, human face has some specific characteristics, such as prior information (including facial landmarks, facial parsing maps, and facial heatmaps), which are ignored by general FSR methods. Therefore, to recover facial images with a much clearer facial structure, researchers begin to develop prior-guided FSR methods.

\begin{figure*}[t]
    \centering
    \includegraphics[width=\linewidth]{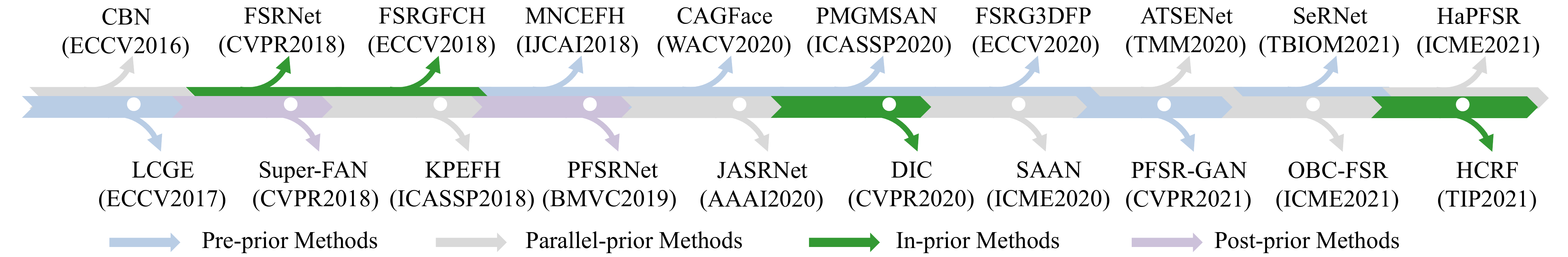}
    \caption{Milestones of prior-guided FSR methods. We simply list their names and venues.}
    \label{prior_guide}
\end{figure*}
Prior-guided FSR methods refer to extracting facial prior information and utilizing it to facilitate face reconstruction. Considering the order of prior information extraction and FSR, we further divide the prior-guided FSR methods into four parts: i) pre-prior methods that extract prior information followed by FSR; ii) parallel-prior methods that extract prior information and FSR simultaneously; iii) in-prior methods that extract prior information from the intermediate results or features at the middle stag, and iv) post-prior methods that extract prior information from FSR results. We illustrate the main frameworks of the four categories in Fig.~\ref{prior1}, outline the development of prior-guided FSR methods in Fig.~\ref{prior_guide} and compare them on several key features in Table~\ref{tab:prior}.

\subsubsection{Pre-prior Methods}\label{prior_loss}
These methods first extract face structure prior information and then feed the prior information to the beginning of FSR model. That is, they always extract prior information from LR face images by an extraction network which can be a pretrained network or a subnetwork associated with the FSR model, then take advantage of the prior information to facilitate FSR. To extract the accurate face structure prior, prior-based loss is always used in these methods to train their prior extraction network, which is defined as:
\begin{equation}\label{prior_loss_function}
\mathcal{L}_{\text{Prior}}= \left\| P_{\text{HR}} - P \right\|_{F},
\end{equation}
where $P_{\text{HR}}$ is the ground truth prior, $P$ is extracted prior from the super-resolved face image, $F$ can be 1 or 2, and the prior can be heatmap, landmark and parsing maps in different methods.

In the early years, both LCGE~\cite{LCGE} and MNCEFH~\cite{MNCEFH} extract landmarks from LR face images to crop the faces into different components, and then predict high-frequency details for different components. However, accurate landmarks are unavailable especially when LR face images are tiny (\emph{i.e.}, 16$\times$16). Thus, researchers turn to facial parsing maps~\cite{PSFR-GAN,SsRNet,CAGFace,PMGMSAN}. PSFR-GAN~\cite{PSFR-GAN}, SeRNet~\cite{SsRNet} and CAGFace~\cite{CAGFace} all pretrain a face structure prior extraction network to extract facial parsing maps. Then all of them except SeRNet directly concatenate the prior and LR face images as the input of the super-resolution model while SeRNet designs its improved residual block (IRB) to fuse the prior and features from LR face images. In addition, PSFR-GAN designs a semantic aware style loss to calculate the gram matrix loss for each semantic region separately. Later on, super-resolution guided by 3D facial priors (FSRG3DFP)~\cite{FSRG3DFP} estimates 3D priors instead of 2D priors to learn 3D facial details and capture facial component information by the spatial feature transform block (SFT).

\setlength{\tabcolsep}{1.00pt}
\begin{table*}
\centering

\caption{Comparison of prior-guided FSR methods. To be short, we use \emph{Pre}, \emph{Parallel}, \emph{In}, and \emph{Post} to denote different prior-guided methods.}
\begin{tabularx}{\linewidth}{m{1.5cm}<{\centering}m{2.4cm}<{\centering}m{5cm}<{\centering}m{1.4cm}<{\centering}X<{\centering}}
\toprule
\multirow{2}{*}{} & \multirow{2}{*}{Methods}   & \multirow{2}{*}{Prior} & \multirow{2}{*}{Extraction} &
\multirow{2}{*}{Fusion Strategies} \\
                  &   &      &  &  \\
\midrule

  \multirow{5}{*}{Pre}&LCGE~\cite{LCGE} &  Landmark &Pretrained & Crop \\ 
  &MNCEFH~\cite{MNCEFH}&Landmark &Pretrained& Crop \\ 
    &PSFR-GAN \cite{PSFR-GAN}  & Parsing map &Pretrained& Concatenation \\
 &CAGFace~\cite{CAGFace} & Parsing map &Pretrained& Concatenation \\ 
  &FSRG3DFP~\cite{FSRG3DFP} &3D prior &Joint& SFT \\ 
  &SeRNet~\cite{SsRNet}&Parsing map&Pretrained&IRB\\
  \hline
\multirow{4}{*}{Parallel}  &CBN~\cite{CBN} &Dense correspondence field & Joint &Concatenation \\
&KPEFH~\cite{KPEFH} &  Parsing map& Joint&$\mathcal{L}_{\text{Parsing}}$ \\ 
&JASRNet~\cite{JASRNet} &  Heatmap& Joint&$\mathcal{L}_{\text{Heatmap}}$ \\ 
&ATSENet~\cite{ATENet1} & Facial boundary heatmap& Joint& FFU \\

\hline
 \multirow{3}{*}{In} &FSRNet~\cite{FSRNet} & Landmark, parsing map, heatmap &Joint &Concatenation \\ 
  &FSRGFCH~\cite{FSRFCH} & Heatmap &Joint&Concatenation \\ 
   &DIC~\cite{DIC}  & Heatmap & Joint& $\mathcal{L}_{\text{Heatmap}}$, AFM\\

   \hline
 \multirow{2}{*}{Post}&Super-FAN~\cite{SuperFAN}  &Heatmap & Joint& $\mathcal{L}_{\text{Heatmap}}$ \\
  &PFSRNet~\cite{PFSRNet}  &  Heatmap&Pretrained& $\mathcal{L}_{\text{Heatmap}}$, $\mathcal{L}_{\text{Attention}}$\\
\bottomrule
\end{tabularx}
\label{tab:prior}
\end{table*}

\subsubsection{Parallel-prior Methods} The above methods ignore the correlation between face structure prior estimation and FSR task: face prior estimation benefits from the enhancement of FSR and vice versa. Thus, parallel-prior methods that perform prior estimation and super-resolution in parallel are proposed, including cascaded bi-network (CBN)~\cite{CBN}, KPEFH~\cite{KPEFH}, JASRNet~\cite{JASRNet}, SAAN~\cite{SAAN}, HaPFSR~\cite{wangicme}, OBC-FSR~\cite{icme_organ} and ATSENet~\cite{ATENet1}. They train the prior estimation and super-resolution networks jointly and require ground truth prior to calculate prior-based loss like Eq. (\ref{prior_loss_function}). 

One of the most representative parallel-prior methods is JASRNet. Specifically, JASRNet utilizes a shared encoder to extract features for super-resolution and prior estimation simultaneously. Through this design, the shared encoder can extract the most expressive information for both tasks. In contrast to JASRNet, ATSENet not only extracts shared features for the two tasks, but also feeds features from the prior estimation branch into the feature fusion unit (FFU) in the super-resolution branch.

\subsubsection{In-prior Methods} Pre- and parallel-prior methods directly extract structure prior information from LR face images. Due to the low-quality of LR face images, extracting accurate prior information is challenging. To reduce the difficulty and improve the accuracy of prior estimation, researchers first coarsely recover LR face images and then extract prior information from the enhanced results of LR face images, including FSRNet~\cite{FSRNet}, FSR guided by facial component heatmaps (FSRGFCH)~\cite{FSRFCH}, HCFR~\cite{hcrf}, deep-iterative-collaboration (DIC)~\cite{DIC} and \cite{DNN,MSFSR,WANG2021106987,LIU2021357,dipfsr}. Similar to parallel-prior methods, in-prior methods always jointly optimize the networks for two tasks.

Specifically, FSRNet~\cite{FSRNet}, FSRGFCH~\cite{FSRFCH} and HCFR~\cite{hcrf} first upsample the LR face images to obtain intermediate results, then extract face structure prior from the intermediate results, and finally make use of the prior and intermediate results to recover the final results. FSRNet and FSRGFCH concatenate the intermediate results and the prior and feed the concatenated results into the following network to recover final SR results while HCFR utilizes the prior to segment the intermediate results and recovers final SR results by random forests. Considering that FSR and prior extraction should facilitate each other, DIC~\cite{DIC} proposes to iteratively perform super-resolution and prior extraction tasks. In the first iteration, DIC recovers a face $I_{\text{SR}^{\text{1}}}$ with super-resolution model and extracts prior (heatmaps) $P_{\text{1}}$ from $I_{\text{SR}^{\text{1}}}$. In the $i$-th iteration, both the LR face image and $P_{i-1}$ are fed into the super-resolution model to obtain $I_{\text{SR}^{i}}$, and then $P_{i}$ can be extracted. In this way, the two tasks can promote each other. Moreover, DIC builds an attention fusion module (AFM) to fuse facial prior and the LR face image efficiently.

\subsubsection{Post-prior Methods}
In contrast to the above methods, post-prior methods extract the face structure prior from SR result rather than LR face image or intermediate result, and utilize the prior to design loss functions, including Super-FAN~\cite{SuperFAN}, progressive FSR network (PFSRNet)~\cite{PFSRNet}, and \cite{gesgnet}. Super-FAN~\cite{SuperFAN} and PFSRNet~\cite{PFSRNet} first super-resolve LR face images and obtain SR results, and then develops a prior estimation network to extract the heatmaps of SR face images and HR ones, and constrains the heatmaps of SR face images and HR ones to be close. PFSRNet further generates multi-scale super-resolved results and applies prior-based loss at every scale. In addition, PFSRNet utilizes heatmaps to generate a mask and calculates facial attention loss $\mathcal{L}_{\text{Attention}}$ based on the masked SR and HR face images. Compared with the above methods, post-prior methods do not require prior extraction during the inference.

\subsubsection{Discussion} All prior-guided FSR methods need the ground truth of the face structure prior to calculate loss in the training phase. During the testing phase, all prior-guided FSR methods except post-prior methods need to estimate the prior. Due to the loss of information caused by image degradation, LR face images increase the difficulty and limit the accuracy of prior extraction in pre-prior methods, further limiting the super-resolution performance. Although parallel-prior methods can facilitate prior extraction and super-resolution simultaneously by sharing feature extraction, the improvement is still limited. In-prior methods extract prior from the intermediate result, which can improve the performance but increase the memory and computation cost caused by iterative super-resolution procedure especially in the iterative method (DIC~\cite{DIC}). In post-prior methods, the prior only plays the role of the supervisor during training, while not participating in inference, and they cannot make full use of the specific prior of the input LR face image. Thus, a method that can exploit the prior fully without increasing additional memory or computation cost is on demand.

\subsection{Attribute-constrained FSR}\label{attribute-constrained}
Facial attribute is also usually exploited in FSR, and they are called attribute-constrained FSR. As a kind of semantic information, facial attribute provides semantic knowledge, \emph{e.g.}, whether people wear glasses, which is useful for FSR. In the following, we will introduce some attribute-constrained FSR methods.

\begin{figure*}[t]
    \centering
    \includegraphics[width=\linewidth]{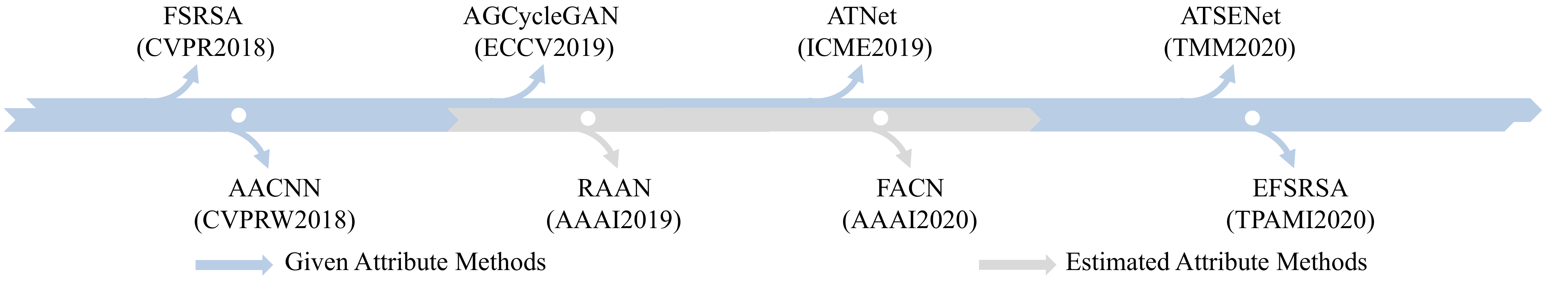}
    \caption{Milestones of attribute-constrained FSR methods. Their names and venues are listed.}
    \label{attribute_constrain}
\end{figure*}

Different from face structure prior information of which acquisition relies on the image itself, attribute information can be available without LR face images, such as in criminal cases where attribute information may not be clear in LR face images but accurately known by witnesses. Thus, some researchers construct networks on the condition that attribute information is given, while others relax this by estimating attributes. According to this concept, attribute-constrained FSR methods can be divided into two frameworks: given attribute methods and estimated attribute methods. The overview is provided in Fig.~\ref{attribute_constrain} and Table~\ref{AGSR}.

\subsubsection{Given Attribute Methods} Given the attribute information, how to integrate it into the super-resolution model is the key. For this problem, attribute-guided conditional CycleGAN (AGCycleGAN)~\cite{AGCycleGAN}, FSR with supplementary attributes (FSRSA)~\cite{FaceAttr}, expansive FSR with supplementary attributes (EFSRSA), attribute transfer network (ATNet)~\cite{ATENet} and ATSENet~\cite{ATENet1}  all directly concatenate attribute information and LR face image (or features extracted from LR face image). AGCycleGAN and FSRSA also feed the attribute into their discriminators to force the super-resolution model to notice the attribute information and develop attribute-based loss to achieve attribute matching, which is defined as:
\begin{equation}
\mathcal{L}_{\text{Attribute}_{D}}=-\log D(I_{\text{HR}}, A) -\log \left(1-D(I_{\text{SR}}, A)\right)-\log \left(1-D(I_{\text{HR}}, \tilde{A})\right),
\end{equation}
where $A$ is attribute matched with $I_{\text{HR}}$ while $\tilde{A}$ is the mismatched one. ATSENet feeds the super-resolved result into an attribute analysis network to calculate attribute prediction loss,
\begin{equation}\label{attribute_based_loss}
    \mathcal{L}_{\text{Attribute}}=\left\|A_{\text{P}}-A_{\text{HR}}\right\|_{F},
\end{equation}
where $A_{\text{P}}$ is the predicted attribute of the network and $A_{\text{HR}}$ is the ground truth attribute.
However, Lee \emph{et al}.~\cite{AACNN} hold that LR face image and attributes belong to different domains, and direct concatenation is unsuitable and may decrease the performance. With regard to this view, Lee \emph{et al}. construct an attribute augmented convolutional neural network (AACNN)~\cite{AACNN}, which extracts features from the attribute to boost face super-resolution.

\begin{table*}

\centering
\caption{Comparison of attribute-constrained FSR methods. "NG" denotes that the information is not given.}
\begin{tabularx}{\linewidth}{X<{\centering}X<{\centering}X<{\centering}m{6cm}<{\centering}}
\toprule
& \multirow{2}{*}{Methods}  & \multirow{2}{*}{\#Attribute}   & \multirow{2}{*}{Attribute embedding methods} \\
                &   &          \\
\midrule
 \multirow{6}{*}{\shortstack{Given }}
 &FSRSA~\cite{FaceAttr}& 18 & Concatenation and $\mathcal{L}_{\text{Attribute}_{D}} $\\&EFSRSA~\cite{FaceAttr1}& 18  & Concatenation and $\mathcal{L}_{\text{Attribute}_{D}} $\\ 
 &AGCycleGAN~\cite{AGCycleGAN} &18   &  Concatenation and $\mathcal{L}_{\text{Attribute}_{D}} $\\ 
&AACNN~\cite{AACNN}   & 38  &  Concatenation \\

&ATNet~\cite{ATENet}   & NG  &  Concatenation and $\mathcal{L}_{\text{Attribute}}$ \\
&ATSENet~\cite{ATENet1}   & NG  &   Concatenation and $\mathcal{L}_{\text{Attribute}}$ \\
\hline
\multirow{2}*{\shortstack{Estimated}} &RAAN~\cite{RAAN} &NG & Attribute channel attention and $\mathcal{L}_{\text{Attribute}}$\\ 
 &FACN~\cite{FACN}    & 18  & Attribute attention mask and $\mathcal{L}_{\text{Attribute}}$\\

\bottomrule
\end{tabularx}
\label{AGSR}
\end{table*}

\subsubsection{Estimated Attribute Methods}
The above-mentioned given attribute methods work on the condition that all attributes are given, making them limited in real-world scenes where some attributes are missing. Although the missed attributes can be set as unknown, such as 0 or random values, the performance may drop sharply. To this end, researchers build modules to estimate attribute information for FSR. In estimated attribute methods, attribute-based loss forces the network to predict attribute information correctly, which is similar to Eq. (\ref{attribute_based_loss}). Estimated attribute methods include residual attribute attention network (RAAN)~\cite{RAAN} and facial attribute capsule network (FACN)~\cite{FACN}. RAAN is based on cascaded residual attribute attention blocks (RAAB). RAAB builds three branches to generate shape, texture, and attribute information, respectively, and introduces two attribute channel attention applied to shape and texture information. In contrast, FACN~\cite{FACN} integrates attributes in capsules. Specifically, FACN encodes LR face image into encoded features, and the features are fed into a capsule generation block that produces semantic capsules, probabilistic capsules, and facial attributes. Then, the attribute is viewed as a kind of mask to refine other features by multiplication or summation. With the combination of three information as input, the decoder of FACN can well recover the final SR results.

\subsubsection{Discussion} Given attribute methods require attribute information, making them only applicable in some restricted scenes. Although the attribute can be set as unknown in these methods, the performance may drop sharply. Towards the estimated attribute methods, they need to estimate the attribute and then utilize the attribute. Compared with given attribute methods, they have a wider range of applications but the accuracy of attribute estimation is difficult to guarantee in practice.

\subsection{Identity-preserving FSR}
\label{sec:identity}
\begin{figure}[t]
    \centering
    \includegraphics[width=\linewidth]{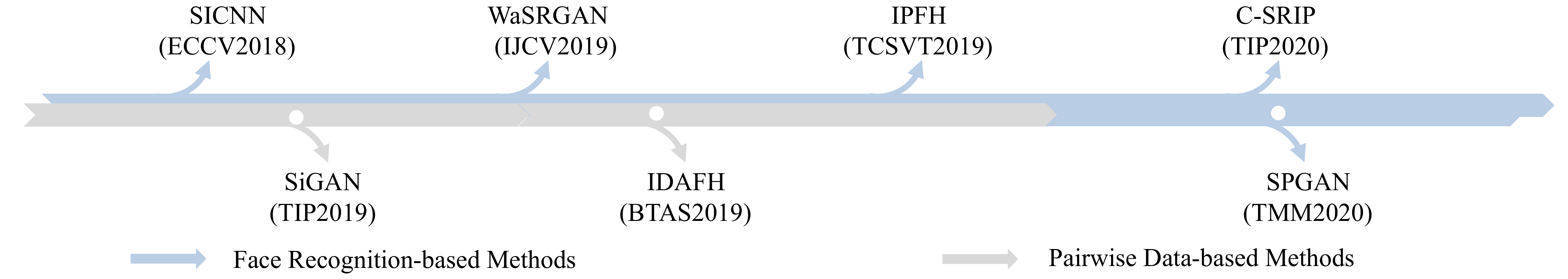}
    \caption{Milestones of identity-preserving FSR methods. Their names and venues are listed.}
    \label{identity_preserving}
\end{figure}

Compared with face structure prior and attribute information, identity information containing identity-aware details is essential and identity-preserving FSR methods have received an increasing amount of attention in recent years. They aim to maintain the identity consistency between SR face image and LR one and improve the performance of down-stream face recognition. We show the overview and comparison of some representative methods in Fig.~\ref{identity_preserving} and Table~\ref{IPSR}.

\subsubsection{Face Recognition-based Methods:}\label{identity_loss_fr}
To maintain identity consistency between $I_{\text{SR}}$ and $I_{\text{HR}}$, in the training phase, a commonly used design is utilizing face recognition network to define identity loss, \emph{e.g.}, super-identity convolutional neural network (SICNN) \cite{SICNN}, face hallucination generative adversarial network (FH-GAN)~\cite{FHGAN}, WaSRGAN~\cite{WaveletSRNet1}, ~\cite{LRFRIPFH}, identity preserving face hallucination (IPFH)~\cite{DRLIPFH}, cascaded super-resolution and identity priors (C-SRIP)~\cite{C-SRIP}, \cite{FRSRResNet,dual_identity,KIM202111,ataer2019verification} and ATSENet~\cite{ATENet1}. The framework of these methods consists of two main components: a super-resolution model, and a pretrained face recognition network (FRN), probably an additional discriminator. The super-resolution model super-resolves the input LR face image, generating $I_{\text{SR}}$ which is fed into FRN to obtain its identity features. Simultaneously, $I_{\text{HR}}$ is also fed into FRN, obtaining its identity features. The identity loss is calculated by
\begin{equation}\label{identity_loss_de}
    \mathcal{L}_{\text{Identity}}=\left\|\text{FR}(I_{\text{HR}})-\text{FR}(I_{\text{SR}})\right\|_{F},
\end{equation}
where FR is the function of FRN. $F$ is 1 in WaSRGAN~\cite{WaveletSRNet1} and 2 in FH-GAN~\cite{FHGAN} and~\cite{FRSRResNet}. Some methods calculate the loss on normalized features~\cite{SICNN,SPLia}, and some use A-softmax loss~\cite{SphereFace,DRLIPFH}. Rather than directly extracting identity features from $I_{\text{SR}}$ and $I_{\text{HR}}$, C-SRIP~\cite{C-SRIP} feeds residual maps between $I_{\text{HR}}$ (or $I_{\text{SR}}$) and $I_{\text{LR}}^{\uparrow_{s}}$ (upsampled by bicubic interpolation), respectively, into FRN, and applies cross-entropy loss on them. Moreover, C-SRIP generates multi-scale face images which are fed into different scale face recognition networks.

To fully explore the identity prior, SPGAN~\cite{SPGAN} feeds identity information extracted by the pretrained FRN to the discriminator at different scales, and designs attention-based identity loss. Firstly, SPGAN generates two attention maps $M_{\text{G}}$ and $M_{\text{D}}$,
\begin{equation}
    E =\mathcal{D}(I_{\text{LR}}, I_{\text{HR}})- \mathcal{D}(I_{\text{LR}},I_{\text{SR}}),
\end{equation}
\begin{equation}
M_{\text{D}}=-\min \left(0, E-\left\|I_{\text{HR}}-I_{\text{SR}}\right\|_2\right),
\end{equation}
\begin{equation}
M_{\text{G}} =\alpha * E+b,
\end{equation}
where $E$ denotes the difference, $*$ denotes the element-wise multiplication, $b$ is identity matrix, and $\alpha$ is a 0-1 matrix. At $i$-th row and $j$-th column, $\alpha_{i,j}$ is 0 when $E_{i,j}$ is negative, otherwise $\alpha_{i,j}$ is 1. Then two attention maps are applied to the identity loss $\mathcal{L}_{\text{Identity}}$,
\begin{equation}
    \mathcal{L}_{\text{Identity}_{\text{SP}}}=\mathcal{L}_{\text{Identity}}*M_{\text{G}}+\mathcal{L}_{\text{Identity}}*M_{\text{D}},
\end{equation}
where $\mathcal{L}_{\text{Identity}_{\text{SP}}}$ is the identity loss of SPGAN.
\begin{table*}
\centering
\caption{Comparison of identity-preserving FSR methods. Notably, $I_{\text{RH}}$ ($I_{\text{RS}}$) is the residual map between $I_{\text{HR}}$ ($I_{\text{SR}}$) and $I_{\text{LR}}^{\uparrow_{s}}$. }
\begin{tabularx}{\linewidth}{m{3.6cm}<{\centering}m{2.6cm}<{\centering}X<{\centering}}
\toprule
\multirow{2}{*}{} & \multirow{2}{*}{Methods}   & \multirow{2}{*}{Loss Functions} \\
                &   &            \\
\midrule

 \multirow{6}{*}{Face Recognition-based} &SICNN~\cite{SICNN} & MSE loss on normalized $\text{FR}(I_{\text{SR}})$ and $\text{FR}(I_{\text{HR}})$  \\ 
 &FH-GAN~\cite{FHGAN}  & \multirow{1}{*}{MSE loss on $\text{FR}(I_{\text{SR}})$ and $\text{FR}(I_{\text{HR}})$} \\ 

&WaSRGAN~\cite{WaveletSRNet1}   &\multirow{1}{*}{$\mathcal{L}_{1}$ loss on $\text{FR}(I_{\text{SR}})$ and $\text{FR}(I_{\text{HR}})$}  \\ 
&C-SRIP~\cite{C-SRIP}    & Cross entropy loss on $\text{FR}(I_{\text{RS}})$ and $\text{FR}(I_{\text{RH}})$    \\ 
&IPFH~\cite{DRLIPFH}    & A-softmax loss on     $\text{FR}(I_{\text{SR}})$ and $\text{FR}(I_{\text{HR}})$    \\ 
 &\multirow{1}*{SPGAN~\cite{SPGAN}}    & \multirow{1}*{Attention-based loss $\mathcal{L}_{\text{Identity}_{\text{SP}}}$} \\
\hline
\multirow{2}*{\shortstack{Pairwise Data-based}} &SiGAN~\cite{SIGAN} & Pair contrastive loss $\mathcal{L}_{\text{Contrastive}}$\\
 &IADFH~\cite{IADFH} & Adversarial face veri?cation loss $\mathcal{L}_{\text{AFVL}}$\\
\bottomrule
\end{tabularx}

\label{IPSR}
\end{table*}

\subsubsection{Pairwise Data-based Methods:}
The training of FRN needs well-labeled datasets. However, a large well-labeled dataset is very costly. One solution is based only on the weakly-labeled datasets. In consideration of this, siamese generative adversarial network (SiGAN)~\cite{SIGAN} takes advantage of the weak pairwise label (in which different LR face images correspond to different identities) to achieve identity preservation. Specifically, SiGAN has twin GANs ($G_1$ and $G_2$) that share the same architecture but super-resolve different LR face images ($I_{\text{LR}}^1$ and $I_{\text{LR}}^2$) at the same time. As the identities of different LR face images are different, the identities of SR results corresponding to LR face images are also varied. Based on this observation, SiGAN designs an identity-preserving contrastive loss that minimizes the difference between same-identity pairs and maximizes the difference between different-identity pairs,
\begin{equation}
\mathcal{L}_{\text{Contrastive}}=(1-y)\frac{1}{2}\left[ \text{max}(0, 0.5-E_{\text{w}})\right]^{2}+y\frac{1}{2}(E_{\text{w}})^{2},
\end{equation}
\begin{equation}
E_{\text{w}}=\left\|F_{\text{E}}(I_{\text{LR}}^{\text{1}}), F_{\text{E}}(I_{\text{LR}}^{\text{2}})\right\|_{1},
\end{equation}
where $F_{\text{E}}$ is a function used to extract features from the intermediate layers of the generators, $E_{\text{w}}$ measures the distance between the features of $I_{\text{LR}}^1$ and $I_{\text{LR}}^2$, $y$ is 1 when two LR face images belong to the same identity, and $y$ is 0 when LR face images belong to different identities.

Instead of feeding the pair data into twin generators, identity-aware deep face hallucination (IADFH)~\cite{IADFH} feeds pair data into the discriminator. Its discriminator is a three-way classifier that generates fake, genuine and imposter: i) HR and SR face images with the same or different identities ($\text{pair}_{\text{1}}$ or $\text{pair}_{\text{2}}$) correspond to the fake, which forces the discriminator to distinguish $I_{\text{HR}}$ and $I_{\text{SR}}$; ii) two different HR face images of the same identity ($\text{pair}_{\text{3}}$) correspond to the genuine; iii) two HR face images with different identities ($\text{pair}_{\text{4}}$) correspond to the imposter. The last two pairs force the discriminator to capture the identity feature. In this pattern, the generator can incorporate the identity information. The loss is called adversarial face veri?cation loss (AFVL),
\begin{equation}
\mathcal{L}_{\text{AFVL(D)}}=\log d_{f}(\text{pair}_{1}) + \log d_{f}(\text{pair}_{2})\\
+\log d_{\text{gen}}(\text{pair}_{3})+\log d_{\text{imp}}(\text{pair}_{4}),
\end{equation}
\begin{equation}
\mathcal{L}_{\text{AFVL(G)}}=\log d_{\text{gen}}(\text{pair}_{1}) +\log d_{\text{imp}}(\text{pair}_{2}),
\end{equation}
where $\mathcal{L}_{\text{AFVL(D)}}$ ($\mathcal{L}_{\text{AFVL(G)}}$) is the loss function of the discriminator (generator), and $d_{\text{f}}, d_{\text{gen}}, d_{\text{imp}}$ (can be -1, 1, 0) are the outputs of the discriminator for fake, genuine and imposter pairs.

\subsubsection{Discussion} Face recognition-based methods design identity loss based on face recognition network which is always pretrained. The training of a face recognition network requires well-labeled datasets which are costly. Instead, pairwise data-based methods take advantage of the contrast between different identities and the similarity between the same identity to maintain identity consistency without well-labeled datasets, which has a wider range of applications.
\subsection{Reference FSR}

\begin{figure}[t]
    \centering
    \includegraphics[width=\linewidth]{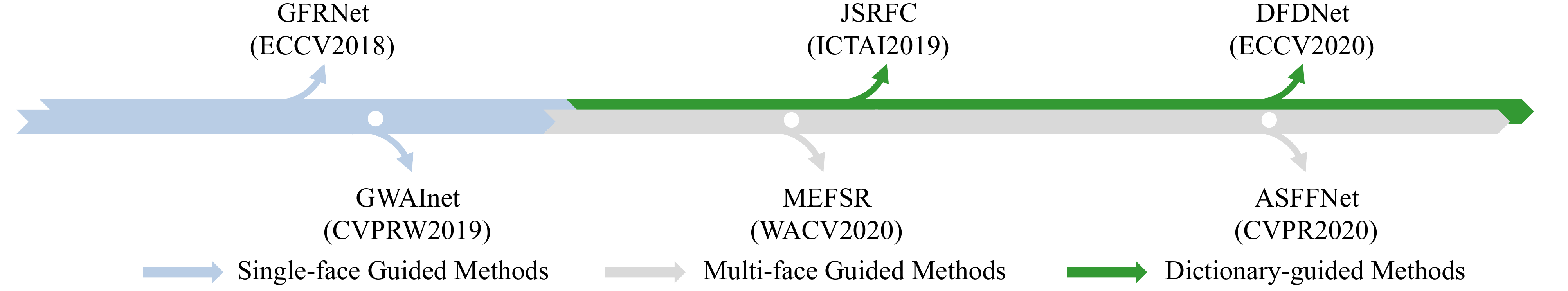}
    \caption{Milestones of reference FSR methods. We simply list their names and venues.}
    \label{face-guide-method}
\end{figure}

The FSR networks discussed all exploit only the input LR face itself. In some conditions, we may obtain the high-quality face image of the same identity of the LR face image, for example, the person of the LR face image may have other high-quality face images. These high-quality face images can provide identity-aware face details for FSR. Thus, reference FSR methods utilize high-quality face image(s) as reference (R) to boost face restoration. Obviously, the reference face image can be only one image or multiple images. According to the number of R, a guided framework can be partitioned into single-face guided, multi-face guided, and dictionary-guided methods.  An overview of reference FSR methods is shown in Fig.~\ref{face-guide-method} and the comparison of them is shown in Table~\ref{GSR}.

\subsubsection{Single-face Guided Methods}
At first, a high-quality face image which shares the same identity with the LR face image serves as R, such as guided face restoration network (GFRNet)~\cite{GFRNet}, GWAInet~\cite{GWAInet}. Since the reference face image and LR face image may have different poses and expressions, which may hinder the recovery of face images, single-face guided methods tend to perform the alignment between the reference face image and the LR face image. After alignment, both the LR face image and aligned reference face image (we name it $I_{\text{w}}$) are fed into a reconstruction network to recover the SR result. The differences between GFRNet and GWAInet include two aspects: i) GFRNet employs landmarks while GWAInet employs flow field to carry out the alignment; ii) in the reconstruction network, GFRNet directly concatenates the LR face image and $I_{\text{w}}$ as the input. Nevertheless, GWAInet builds a GFENet to extract features from $I_{w}$ and transferring useful features of $I_{w}$ to the reconstruction network to recover SR results.
\begin{table*}
\centering
\caption{Comparison of reference FSR methods. "-" denotes that the method does not contain the procedure.}
\begin{tabularx}{\linewidth}{X<{\centering}X<{\centering}m{2cm}<{\centering}X<{\centering}X<{\centering}}
\toprule
 & \multirow{2}{*}{Methods}   & \multirow{2}{*}{Same identity} & \multirow{2}{*}{Alignment} & \multirow{2}{*}{Utilization of R} \\
                &   &        &  &  \\
\midrule
 \multirow{2}{*}{\shortstack{Single-face guided}} &GFRNet~\cite{GFRNet} &\checkmark  & Landmark &Concatenation \\ 
 &GWAInet~\cite{GWAInet} &   \checkmark &Flow field& GFENet \\ 
 \hline
 \multirow{2}{*}{\shortstack{Multi-face guided}}&ASFFNet~\cite{ASFFNet}  &  \checkmark  & Moving least-square  & AFFB \\ 
   &MEFSR~\cite{MEFSR}  & \checkmark & -& PWAve\\ 
         \hline
  \multirow{2}{*}{\shortstack{Dictionary-guided}}   &JSRFC~\cite{JSRFC}  &   $\times$& Landmark & Concatenation\\ 

  &DFDNet~\cite{DFDNet} & $\times$ &- & DFT\\ 
\bottomrule
\end{tabularx}
\label{GSR}
\end{table*}

\subsubsection{Multi-face Guided Methods}
Single-face guided methods set the problem as an LR face image only has one high-quality reference face image, but in some applications many high-quality face images are available, and they can further provide more complementary information for FSR. Adaptive spatial feature fusion network (ASFFNet)~\cite{ASFFNet} is the first to explore multi-face guided FSR. Given multiple reference images, ASFFNet first selects the best reference image which should have the most similar pose and expression with LR face image by guidance selection module. However, misalignment and illumination differences still exist in the reference face image and the LR face image. Thus, ASFFNet applies weighted least-square alignment~\cite{MLS} and AdaIN~\cite{AdaIn} to cope with these two problems. Finally, they design an adaptive feature fusion block (AFFB) to generate an attention mask that is used to complement the information from LR face image and R. Multiple exemplar FSR (MEFSR)~\cite{MEFSR} directly feed all reference faces into weighted pixel average (PWAve) module to extract information for face restoration.

\subsubsection{Dictionary-guided Methods}

It is observed that different people may have similar facial components. According to this observation, dictionary-guided methods are proposed, including joint super-resolution and face composite (JSRFC)~\cite{JSRFC} and deep face dictionary network (DFDNet)~\cite{DFDNet}. Dictionary-guided methods do not require the identity consistency between the reference face image and the LR face image, but build a component dictionary to boost face restoration. For example, JSRFC selects reference images which have similar components with the LR face image (every reference face image is labeled with a vector to indicate which components are similar.). Then, it aligns LR face image with the reference face image and extracts the corresponding components as a component dictionary. Finally, the dictionary components are used for the following face restoration. Different from JSRFC, Li \emph{et al}.~\cite{DFDNet} build multi-scale component dictionaries based on features of the entire dataset. They use pretrained VGGFace~\cite{VGGFace2} to extract features in different scales from high-quality faces, and then crop and resample four components with landmarks, and then cluster obtain K classes for every component by K-means. Given component dictionaries, they first select the most similar atoms for every component by the inner product, and then transfer the features from dictionary to the LR face image by dictionary feature transfer (DFT).

\subsubsection{Discussion} Single-face and multi-face guided FSR methods require one or multiple additional high-quality face image(s) with the same identity as the LR face image, which facilitates face restoration but limits their application since the reference image may not exist. In addition, the alignment between low-quality LR face image and high-quality reference face image is also challenging in the reference FSR. Dictionary-guided methods break the restriction of the same identity, broadening the application but increasing the difficulty of face reconstruction.

\subsection{Experiments and Analysis}
To have a clear view of deep learning-based FSR methods, we compare the PSNR, SSIM and LPIPS performance of the state-of-the-art algorithms on commonly used benchmark datasets (including CelebA~\cite{celeba}, VGGFace2~\cite{VGGFace2} and CASIA-WebFace~\cite{CASIA-WebFace}) with upscale $\times$4, $\times$8 and $\times$16. Considering that the reference FSR methods are different from other FSR methods, we compare other FSR methods and reference FSR methods individually.


\subsubsection{Comparison Results of FSR Methods} We first introduce the experimental settings and analyze the results of FSR methods.

\textbf{Experimental Setting:} For CelebA~\cite{celeba} dataset, 168,854 images are used for training and 1,000 images for testing following DIC~\cite{DIC}. All the images are cropped and resized into 128$\times$128 as $I_{\text{HR}}$. We apply the degradation model in Eq. (\ref{eq3}) to generate $I_{\text{LR}}$. Facial landmarks are detected by~\cite{openface1,openface2,openface3} and heatmaps are generated according to the landmarks. For facial parsing map, we adopt pretrained BiSeNet~\cite{BiSeNet} to extract the parsing map from $I_{\text{HR}}$. For quality evaluation, PSNR and SSIM are introduced and both of them are computed on the Y channel of YCbCr space, which also follows DIC~\cite{DIC}. In addition, we further introduce the LPIPS to evaluate the performance of all comparison approaches. For the optimizer and learning rate when retraining different methods, we follow the setting in their original papers.

\begin{figure*}
	\centering
    \includegraphics[width=\linewidth]{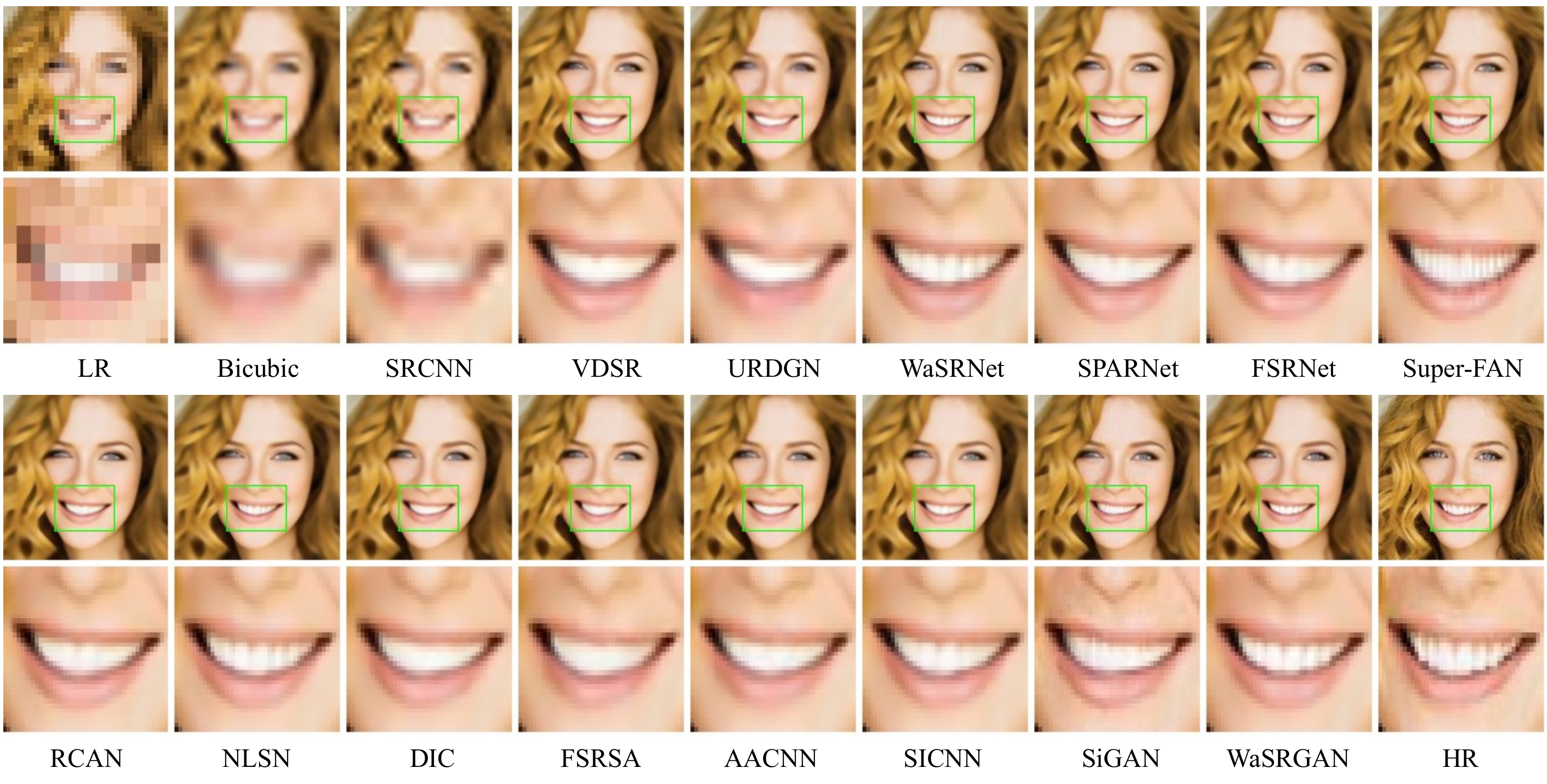}
	\caption{Qualitative comparison of different FSR approaches for $\times$4 super-resolution reconstruction.}
	\label{fig:upscale_x4}
\end{figure*}

\begin{figure*}[t]
	\centering
    \includegraphics[width=\linewidth]{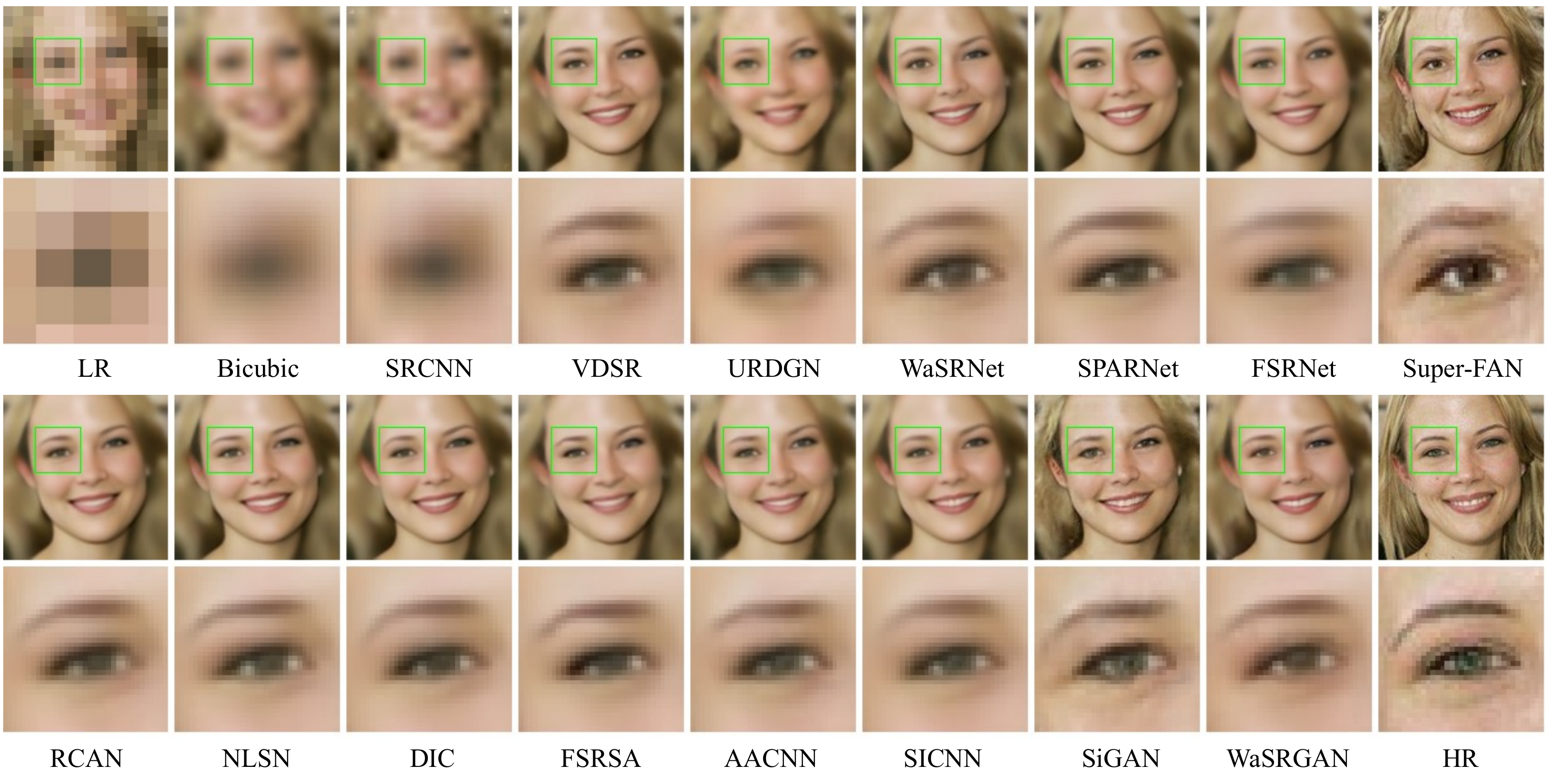}
	\caption{Qualitative comparison of different FSR approaches for $\times$8 super-resolution reconstruction.}
	\label{fig:upscale_x8}
\end{figure*}

\begin{figure*}[t]
	\centering
    \includegraphics[width=\linewidth]{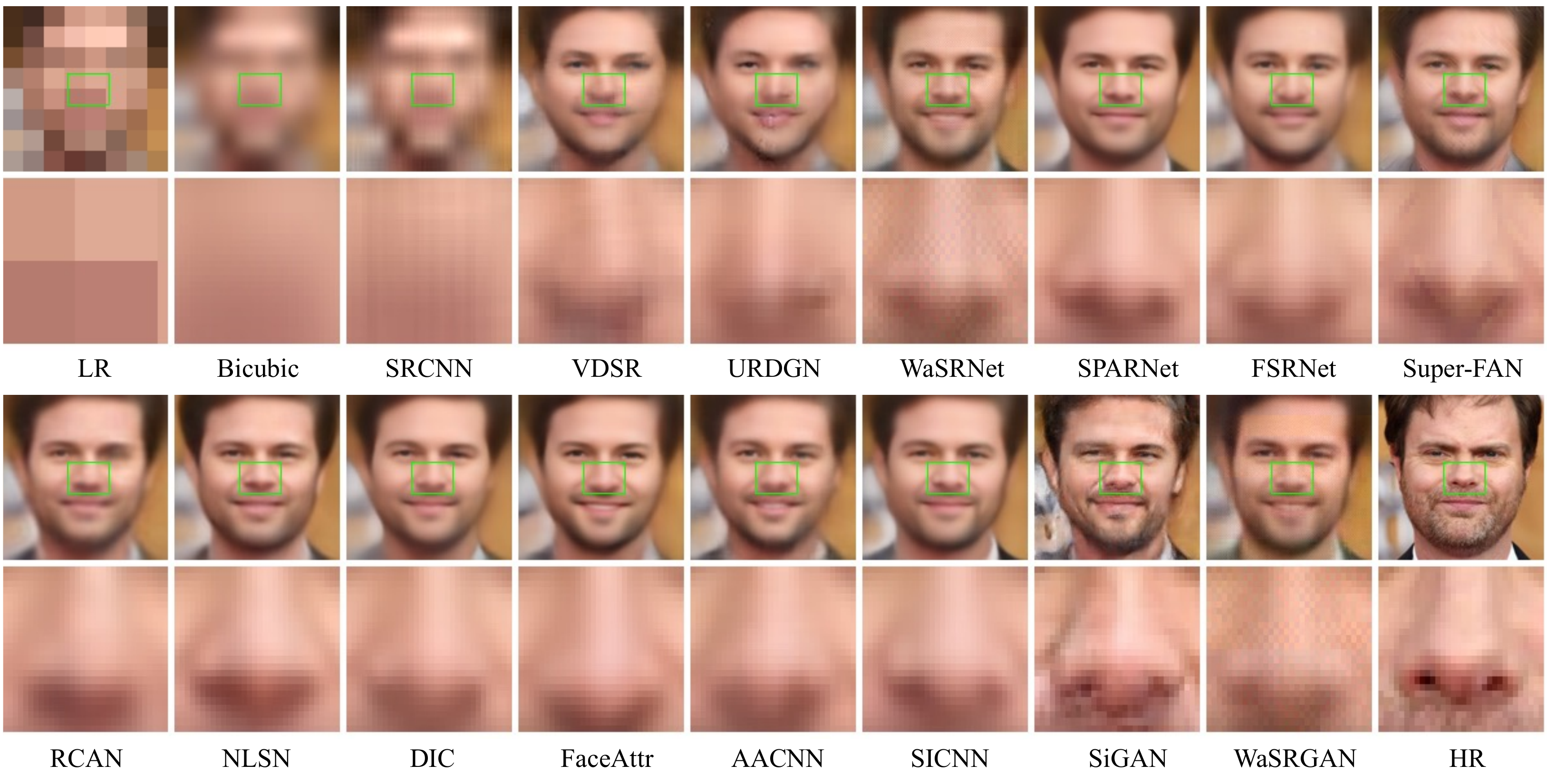}
	\caption{Qualitative comparison of different FSR approaches for $\times$16 super-resolution reconstruction.}
	\label{fig:upscale_x16}
\end{figure*}

\textbf{Experimental Results:} We list and compare the results of some representative FSR methods in Table~\ref{comparison}, including four general image super-resolution methods: super-resolution using deep convolutional networks (SRCNN)~\cite{SRCNN}, VDSR~\cite{VDSR}, residual channel attention network (RCAN)~\cite{RCAN}, non-local sparse network (NLSN)~\cite{nlsn}, three general FSR methods: URDGN~\cite{URDGN}, WaSRNet~\cite{WaveletSRNet}, SPARNet~\cite{SPARNet}, three prior-guided FSR methods: FSRNet~\cite{FSRNet}, Super-FAN~\cite{SuperFAN}, DIC~\cite{DIC}, two attribute-constrained FSR methods: FSRSA~\cite{FaceAttr1}, AACNN~\cite{AACNN}, and three identity-preserving FSR methods: SICNN~\cite{SICNN}, SiGAN~\cite{SIGAN}, and WaSRGAN~\cite{WaveletSRNet1}. Except that, we also report the parameters and FLOPs of these methods in the last two columns of Table \ref{comparison}. Note that the parameter and FLOPs are associated with the model with upscale $\times$8. In addition, we also present the visual comparisons between a few state-of-the-art algorithms in Fig.\ref{fig:upscale_x4}, Fig.\ref{fig:upscale_x8} and Fig.\ref{fig:upscale_x16}.

From these objective metrics and visual comparison results, we have the following observations:

(i) The retrained state-of-the-art (SOTA) general image super-resolution methods, such as RCAN and NLSN, are very competitive and even outperform the best FSR methods in terms of PSNR and SSIM. Meanwhile, as a general FSR method, SPARNet obtains the best performance among all the FSR methods. RCAN, NLSN, and SPARNet all do not explicitly incorporate the prior knowledge of face image, but they have obtained outstanding results. It shows that the design and optimization of the network is very important, and a well-designed network will have stronger fitting capabilities (less reconstruction errors). This observation will enlighten us that when we are designing a FSR deep network, it should be based on a strong backbone network.

(ii) The terms of RCAN* and NLSN* in Table \ref{comparison} represent the pretrained models on general training images, and we directly download these models from the authors' pages. Note that the pretrained results under certain magnification factors are not given (indicated as `-' in the table) because these methods are not trained under these magnification factors. RCAN and NLSN achieve better performance than RCAN* and NLSN*. This demonstrates that models trained by general images are not suitable for FSR but general image super-resolution methods trained by face images may perform well (sometimes even better than FSR methods on face images). Therefore, if we want to know and compare the performance of a newly proposed general image super-resolution on the task of FSR, we cannot directly use the pretrained model released by the authors, but should retrain the model on the face image dataset. It should be noted that the objective results of these GAN-based FSR methods (\emph{e.g.}, URDGN, FSRSA, SiGAN and WaSRGAN) are worse than those of NLSN*. This is mainly because that they often cannot get a better MSE due to the introduction of adversarial losses, which tend to allow the models to obtain perceptually better SR results but large reconstruction errors.

(iii) Compared with general image super-resolution methods and general FSR methods, these methods that incorporate facial characteristics do not perform well in terms of PSNR and SSIM. Nevertheless, we cannot conclude that it is meaningless to develop FSR methods that use facial characteristics. This is mainly because PSNR and SSIM may be not good assessment metrics for the task of image super-resolution \cite{lpips}, let alone for the task of FSR, in which human perception will be more important. To further exploit the super-resolution reconstruction capacity, we also introduce another assessment metric, LPIPS, which is more in line with human judgement. From the LPIPS results, we learn that these methods with low PSNR and SSIM may produce very good performance in terms of LPIPS, please refer to Super-FAN and SiGAN. This indicates that these methods that introduce facial characteristics can well represent the face image and recover the face contours and discriminant details.

(iv) When we compare FSR methods that use different facial characteristics, such as face structure prior, attributes, and identity, it is difficult to say which type of characteristic is more effective for FSR. Because these methods often use different backbone networks, and it is difficult to determine whether their performance changes are caused by the difference in the backbone network itself or because of the introduction of different facial characteristics. In practice, we can first develop a strong backbone and then incorporate facial characteristics to boost FSR.

\begin{table}\setlength{\tabcolsep}{4pt}

  \caption{Quantitative evaluation of various FSR methods on CelebA, in terms of PSNR, SSIM and LPIPS for $\times$4, $\times$8 and $\times$16. The \textcolor{red}{best}, the \textcolor{blue}{second-best} and the \underline{third-best} results are emphasized with red, blue and underscore respectively. Note that Params and FLOPs are calculated for $\times$8 super-resolution model.}

  \small
   \begin{tabularx}{\linewidth}{m{2.3cm}<{\centering}X<{\centering}X<{\centering}   X<{\centering} X<{\centering}X<{\centering}  X<{\centering} X<{\centering}  X<{\centering}  X<{\centering}X<{\centering}  X<{\centering}}
    \hline
        \multirow{2}{*}{Methods} &  \multicolumn{3}{c}{$\times$4} &\multicolumn{3}{c}{$\times$8} & \multicolumn{3}{c}{$\times$16} &\multirow{2}{*}{Params}&\multirow{2}{*}{FLOPs}\\ \cline{2-10}
         &PSNR$\uparrow$&SSIM$\uparrow$&LPIPS$\downarrow$ &PSNR$\uparrow$ &SSIM$\uparrow$&LPIPS$\downarrow$ & PSNR$\uparrow$&SSIM$\uparrow$&LPIPS$\downarrow$& &\\
     \hline
   \multicolumn{12}{c}{General Image Super-Resolution Methods}\\
  \hline
  SRCNN~\cite{SRCNN}&28.04&0.837&0.160 & 23.93 & 0.635 &0.256 &20.54&0.467 & 0.291&0.01M&0.3G\\
  VDSR~\cite{VDSR}&31.25 &0.906 &0.055 & 26.36&0.761 &0.112 &22.42 &0.594 &0.186 & 0.6M &11.0G\\
   RCAN~\cite{RCAN}&\underline{31.69}&\textcolor{blue}{0.913}&0.051 &27.30&\underline{0.799}&0.100&23.32&0.641&0.204&15.0M&4.7G\\
    RCAN*~\cite{RCAN}&26.30&0.769&0.177 &22.17&0.521&0.265&-&-&-&15.0M&4.7G\\
    NLSN~\cite{nlsn}& \textcolor{red}{32.08}& \textcolor{red}{0.919} & \underline{0.044}& \textcolor{red}{27.45}&\textcolor{red}{0.804} & 0.091& \textcolor{red}{23.69}& \textcolor{blue}{0.671}& 0.154&43.4M& 22.9G\\
    NLSN*~\cite{nlsn}&30.82 &0.899& 0.065& -& -& -&- &- &-&43.4M&22.9G\\
   \hline
      \multicolumn{12}{c}{General FSR Methods}\\
  \hline
   URDGN~\cite{URDGN}&30.11&0.884&0.075&25.62&0.726&0.148&22.29&0.579&0.185&1.0M&14.6G\\
   WaSRNet~\cite{WaveletSRNet}&30.92&0.908&0.051&26.83&0.787&\underline{0.089}&23.13&0.634&0.160&71.5M&19.2G\\

   SPARNet~\cite{SPARNet}&\textcolor{blue}{31.71}&\textcolor{blue}{0.913}&0.048&\textcolor{blue}{27.44}&\textcolor{red}{0.804}&\underline{0.089}&\textcolor{blue}{23.68}&\textcolor{red}{0.674}&\underline{0.139}&10.0M&7.2G\\
      \hline
      \multicolumn{12}{c}{Prior-guided FSR Methods}\\
  \hline
   FSRNet~\cite{FSRNet}&31.46&0.908&0.052&26.66&0.771&0.110&23.04&0.629&0.175&3.1M&39.0G\\
   Super-FAN~\cite{SuperFAN}&31.17&0.905&\textcolor{blue}{0.040}&27.08&0.788&\textcolor{red}{0.058}&23.42&0.652&\textcolor{blue}{0.125}&1.3M&1.1G\\
   DIC~\cite{DIC}&31.44&0.909&0.053&\underline{27.41}&\textcolor{blue}{0.802}&0.092&23.47&0.657&0.160&20.8M&14.8G\\
         \hline
      \multicolumn{12}{c}{Attribute-constrained FSR Methods}\\
  \hline
   FSRSA~\cite{FaceAttr1}&30.80&0.898&0.058&26.19&0.757&0.111&22.84&0.630&0.153&76.9M&0.9G\\

   AACNN~\cite{AACNN}&31.30&0.907&0.052&26.68&0.773&0.100&22.98&0.626&0.171&3.3M&0.2G\\

        \hline
      \multicolumn{12}{c}{Identity-preserving FSR Methods}\\
  \hline
   SICNN~\cite{SICNN}&31.59&\underline{0.911}&0.050&27.18&0.793&0.095&\underline{23.50}&\underline{0.662}&0.152&4.9M&5.4G\\
   SiGAN~\cite{SIGAN}&30.68&0.892&\textcolor{red}{0.034}&25.63&0.740&\textcolor{blue}{0.062}&22.18&0.596&\textcolor{red}{0.099}&19.5M&5.7G\\
   WaSRGAN~\cite{WaveletSRNet1}&30.72&0.907&0.045&25.55&0.765&0.092&22.78&0.625&0.148&71.5M&19.2G\\

      \hline
   \end{tabularx}
\label{comparison}

  \end{table}


\subsubsection{Comparison Results of Reference FSR Methods} The above FSR methods only require LR face images as input, while the reference FSR methods require LR face images and reference images. It is unfair to directly compare with these methods that do not use auxiliary high-resolution face images. Therefore, we compare the performance of the reference FSR methods individually.

\textbf{Experimental Setting:} Following ASFFNet~\cite{ASFFNet}, VGGFace2~\cite{VGGFace2} is reorganized into 106,000 groups and every group has 3-10 high-quality face images of the same identity, in which 10,000 groups are used for training set, 4,000 groups are for validation set and the remaining are testing set. In addition, two testing sets based on CelebA~\cite{celeba} and CASIA-WebFace~\cite{CASIA-WebFace} are also used, and each set contains 2,000 groups with 3-10 high-quality face images. We utilize facial landmarks to crop and resize all images into 256$\times$256 as high-quality face images. To generate $I_{\text{LR}}$, the degradation model Eq. (\ref{eq5}), where $J$ and $\downarrow$ are embodied as JPEG compression with quality $q$ and bicubic interpolation respectively, is applied to the high-quality images. We consider two types of blur kernels, \emph{i.e.}, Gaussian blur and motion blur kernels, and randomly sample the scale $s$ from \{1:0.1:8\}, the noise level from \{0:1:15\}, and the compression quality factor $q$ from \{10 : 1 : 60\}~\cite{ASFFNet}. PSNR, SSIM and LPIPS~\cite{lpips} are used as metrics.

\textbf{Experimental Results:} The experimental results are shown in Table~\ref{comparison_reference}. To be specific, we list the results of GFRNet~\cite{GFRNet}, GWAInet~\cite{GWAInet} and the latest proposed ASFFNet~\cite{ASFFNet} on CelebA~\cite{celeba}, VGGFace2~\cite{VGGFace2} and CASIA-WebFace~\cite{CASIA-WebFace} with upscale $\times$8. Note that all the results are copied from the paper~\cite{ASFFNet} since we have difficulty in reproducing these methods. Note that GFRNet and GWAInet are single-face guided methods while ASFFNet is multi-face guided method. To be fair, the reference image of GFRNet and GWAInet is the same as the selected image in ASFFNet. From Table \ref{comparison_reference}, it is obvious that multi-face guided method ASFFNet performs better than single-face guided methods (GWAInet and GFRNet). ASFFNet considers the illumination difference between the reference face image and the LR face image, which is ignored by GFRNet and GWAInet, and builds a well-designed AFFB instead of simple concatenation to adaptively the features of the reference face image and the LR face image. These two points contribute to the excellent performance of ASFFNet. Thus, difference (\emph{i.e.}, misalignment, illumination difference, \emph{etc}.) elimination and effective information fusion of the reference face image and the LR face image are both important in reference FSR methods.



\begin{table}
  \caption{Quantitative evaluation of various reference FSR methods on CelebA~\cite{celeba}, VGGFace2~\cite{VGGFace2} and CASIA-WebFace~\cite{CASIA-WebFace}, in terms of PSNR, SSIM and LPIPS for $\times$8. The best, the second-best and the third-best results are emphasized with \textcolor{red}{red}, \textcolor{blue}{blue} and \underline{underscore} respectively.}
  \begin{tabularx}{\linewidth}{m{2.5cm}<{\centering}  X<{\centering}X<{\centering}   X<{\centering} X<{\centering}X<{\centering}  X<{\centering} X<{\centering}  X<{\centering}  X<{\centering}}
    \hline
        \multirow{2}{*}{Methods} &  \multicolumn{3}{c}{CelebA~\cite{celeba}}& \multicolumn{3}{c}{VGGFace2~\cite{VGGFace2}} & \multicolumn{3}{c}{CASIA-WebFace~\cite{CASIA-WebFace}} \\ \cline{2-10}
         & PSNR$\uparrow$&SSIM$\uparrow$&LPIPS$\downarrow$ &PSNR$\uparrow$ &SSIM$\uparrow$& LPIPS$\downarrow$ & PSNR$\uparrow$& SSIM$\uparrow$& LPIPS$\downarrow$\\
     \hline
  GFRNet~\cite{GFRNet} & \textcolor{blue}{25.93} &\textcolor{blue}{0.901} & \underline{0.227}&\underline{23.85} & \textcolor{blue}{0.879} &\underline{0.263}& \textcolor{blue}{27.19} &\textcolor{blue}{0.912}&\underline{0.307}\\
   GWAInet~\cite{GWAInet}&\underline{25.77}&\textcolor{blue}{0.901}&\textcolor{blue}{0.210} &\textcolor{blue}{23.87}&\textcolor{blue}{0.879}&\textcolor{blue}{0.261}&\underline{27.18}&\underline{0.910}&\textcolor{blue}{0.250}\\
   ASFFNet~\cite{ASFFNet}&\textcolor{red}{26.39}&\textcolor{red}{0.905}&\textcolor{red}{0.185}&\textcolor{red}{24.34}&\textcolor{red}{0.881}&\textcolor{red}{0.238}&\textcolor{red}{27.69}&\textcolor{red}{0.921}&\textcolor{red}{0.219}\\
      \hline
  \end{tabularx}
  \label{comparison_reference}
  \end{table}

\subsection{Joint FSR and Other Tasks}
Although the above FSR methods have achieved a breakthrough, FSR is still challenging and complex since the input face images are often affected by many factors, including shadow, occlusion, blur, abnormal illumination, \emph{etc}. To recover these face images efficiently, some work is proposed to consider degradation caused by low-quality and other factors together. Moreover, researchers also jointly perform FSR and other tasks. 
In the following, we will review these joint FSR and other tasks methods.

\subsubsection{Joint Face Completion and Super-Resolution}
Both low-resolution and occlusion or shadowing always coexist in the real-world face images. Thus, the restoration of faces degraded by these two factors is important. The simplest way is to first complete the occluded part and then super-resolve the completed LR face images \cite{OFHNet}. However, the results always contain large artifacts due to the accumulation of errors. Cai \emph{et al}.~\cite{FCSR-GAN2} propose the FCSR-GAN method which pretrains a face completion model (FCM), and combines FCM with super-resolution model (SRM), then trains SRM with the fixed FCM, and finally finetunes the whole network. Then, Liu \emph{et al}.~\cite{MFG-GAN} propose a graph convolution pyramid blocks, which only needs one step to be trained rather than multiple steps of FCSR-GAN. In contrast, Pro-UIGAN~\cite{pro-uigan} utilizes facial landmark to capture facial geometric prior and recovers occluded LR face images progressively.

\subsubsection{Joint Face Deblurring and Super-Resolution}
Blurry LR face images always arise in real surveillance and sports videos, which cannot be recovered effectively by a single task model, \emph{e.g.}, super-resolution or deblurring model. In the literature, Yu \emph{et al}.~\cite{SCGAN} develop SCGAN to deblur and super-resolve the input jointly. Then, Song \emph{et al}.~\cite{FSGN} find that the previous methods ignore the utilization of facial prior information and the recovered face image are lack of high-frequency details. Thus, they first utilize a parsing map and LR face image to recover a basic result, and then feed the basic result into detail enhancement module to compensate high-frequency details from the high-quality exemplar. Later on, DGFAN~\cite{DGFAN} develops two feature extraction modules for different tasks to extract features, and imports them into well-designed gated fusion modules to generate deblurred high-quality results. Xu \emph{et al}. \cite{SRBIP} incorporate face recognition network with face restoration to improve the identi?ability of the recovered face images.

\subsubsection{Joint Illumination Compensation and FSR}
Abnormal illumination FSR has also attracted the attention of many scholars. SeLENet~\cite{Selenet} decomposes a face image into a normal face, an albedo map and a lighting coef?cient, then replaces the lighting coef?cient with the standard ambient white light coef?cient, and then reconstructs the corresponding neutral light face image. Ding \emph{et al}.~\cite{dark} build a pipeline of face detection, and then recover the detected faces with landmarks. Zhang \emph{et al}.~\cite{copyandpaste} utilize a normal illumination external HR guidance to guide abnormal illumination LR face images for illumination compensation. They develop a copy-and-paste GAN (CPGAN), including an internal copy-and-paste network to utilize face intern information for reconstruction, and an external copy-and-paste network is applied to compensate illumination. Based on CPGAN, they further improve the external copy-and-paste network by introducing recursive learning and incorporating landmark estimation and develop the recursive CPGAN~\cite{Re-CPGAN}. In contrast, Yasarla \emph{et al}. \cite{NASFE} introduce network architecture search into face enhancement to design efficient network and extract identity information from HR guidance to restore face images.

\subsubsection{Joint Face Alignment and Super-Resolution}
The above FSR methods require the all the HR training sample to be aligned. Thus, the misalignment of the input LR face image to the training face images often leads to sharp performance decrease and artifacts. Therefore, a set of joint face alignment and super-resolution methods are developed. Yu \emph{et al}.~\cite{TDN} insert multiple spatial transformer networks (STN)~\cite{STN} into the generator to achieve face alignment, and develop TDN and MTDN~\cite{MTDN}. As LR face images can be noisy and unaligned, Yu \emph{et al}. build the TDAE method~\cite{TDAE}. TDAE first upsamples and coarsely aligns LR face images to produce $I_{\text{CSR}}$, then downsamples $I_{\text{CSR}}$ and obtains $I_{\text{CLR}}$ to reduce noise, and then upsamples $I_{\text{CLR}}$ for the final reconstruction.

\subsubsection{Joint Face Frontalization and Super-Resolution}
Faces in the real world have various poses, and some of them may not be frontal. When existing FSR methods are applied to non-frontal faces, the reconstruction performance drops sharply and has poor visual quality. Artifacts exist even when FSR and face frontalization are performed in sequence or inverse order. To alleviate this problem, the method in~\cite{FH1} first takes advantage of STN and CNN to coarsely frontalize and hallucinate the faces, and then designs a fine upsampling network for refining face details. Yu \emph{et al}.~\cite{FH} propose a transformative adversarial neural network for joint face frontalization and hallucination. The method builds a transformer network to encode non-frontal LR face images and frontal LR ones into the latent space and requires the non-frontal one to be close to the frontal one, and then the encoded latent representations are imported into the upsampling network to recover the final results. Tu \emph{et al}.~\cite{MDFR} first train face restoration network and face frontalization network separately, and then propose task-integrated training strategy to merge two networks into a unified network for face frontalization and super-resolution. Note that face alignment aims to generate SR face images with the same pose as HR ones while face frontalization is to recover frontal SR faces from non-frontal LR faces.

\subsection{Related Applications}
Except the above-mentioned FSR methods and joint methods, a large number of new methods related to FSR have emerged in recent years, including face video super-resolution, old photo restoration, audio-guided FSR, 3D FSR, \emph{etc}., which are introduced in the following.

\subsubsection{Face Video Super-Resolution}
Faces usually appear in LR video sequences, such as surveillance. The correlation between frames can provide more complementary details, which benefit the face reconstruction. One direct solution is to fuse multi-frame information and exploit inter-frame dependency~\cite{IGGAN}. The approach of~\cite{MISO} employs a generator to generate the SR results for every frame, and a fusion module is applied to estimate the central frame. Considering that the aforementioned methods cannot model the complex temporal dependency, Xin \emph{et al}.~\cite{MAFN} propose a motion-adaptive feedback cell which captures inter-frame motion information and updates the current frames adaptively. In \cite{SECA}, based on the assumption that multiple super-resolved frames are crucial for the reconstruction of the subsequent frame, and thus it designs a recurrence strategy to make better use of inter-frame information. Inspired by the powerful transformer, the work of \cite{VidFace} develops the first pure transformer-based face video hallucination model. MDVDNet~\cite{MDVDNet} incorporates multiple priors from the video, including speech, semantic elements and facial landmarks to enhance the capability of deep learning-based method.

\subsubsection{Old Photo Restoration}
Restoration of old pictures is vital and difficult in the real world since the degradation is too complex to be stimulated. Naturally, one solution is to learn the mapping from a real LR face image (regarding real old images as real LR face images) to an artificial LR face images, and then apply the existing FSR methods to the generated artificial LR face image. BOPBL~\cite{OPR} proposes to transform images at latent space rather than image space. Specifically, BOPBL first encodes real and artificial LR face images into the same latent space $S_{1}$, and encodes HR face images into another latent space $S_{2}$, and then maps $S_{1}$ into $S_{2}$ by a mapping network.

\subsubsection{Audio-guided FSR}
Considering that audio carries face-related information \cite{oh2019speech2face}, Meishvili \emph{et al}.~\cite{FSA} develop the first audio-guided FSR method. Due to the difference of multi-modal, they build two encoders to encode image and audio information. Then the encoded representations of images and the audio are fused, and the fused results are fed into the generator to recover the final SR results. The introduction of the audio in FSR is novel and inspires researchers to exploit cross modal information, but is challenging due to the differences between different modalities.



\subsubsection{3D FSR}
Human face is the most concerned object in the field of computer vision. With the development of 2D technology, a large number of 3D methods are often proposed because they can provide more useful features for face reconstruction and recognition. In the FSR society, the early 3D FSR approach is proposed by Pan \emph{et al.} \cite{pan2006super}. In \cite{berretti2012superfaces}, Berretti \emph{et al.} propose a superface model from a sequence of low-resolution 3D scans. The approach of \cite{liang20143d} takes only the rough, noisy, and low-resolution depth image as input, and predicts the corresponding high-quality 3D face mesh. By establishing the correspondence between the input LR face and 3D textures, Qu \emph{et al.} present a patch-based 3D FSR on the mesh \cite{qu2017robust}. Benefiting from the development of deep learning technology, most recently, a 3D face point cloud super-resolution network approach is developed to infer the high-resolution data from low-resolution 3D face point cloud data~\cite{li20213d}.

\section{Conclusion and Future Directions}\label{Sec5}
In this review, we have presented a taxonomy of deep learning-based FSR methods. According to facial characteristics, this field can be divided into five categories: general FSR methods, prior-guided FSR methods, attribute-constrained FSR methods, identity-preserving FSR methods, and reference FSR methods. Then, every category is further divided into some subcategories depending on the design of the network architecture or the specific utilization of facial characteristics. In particular, general FSR methods are further divided into basic CNN-based methods, GAN-based methods, reinforcement learning-based methods, and ensemble learning-based methods. Besides, other methods combining facial characteristics are categorized according to the specific utilization pattern of facial characteristics. We also compare the performance of state-of-the-arts and give some deep analysis. Of course, FSR technique is not limited to the methods we presented, and a panoramic view of this fast-expanding field is rather challenging, thereby resulting in possible omissions. Therefore, this review serves as a pedagogical tool, providing researchers with insights into typical methods of FSR. In practice, researchers could use these general guidelines to develop the most suitable technique for their specific studies.

Despite great breakthroughs, FSR still presents many challenges and is expected to continue its rapid growth. In the following, we simply provide an outlook on the problems to be solved and trends to expect in the future.


\textbf{Design of Network.}
From the comparison results with the SOTA general image super-resolution methods, we learn that the backbone network has a crucial impact on the performance, especially in terms of PSNR and SSIM. Therefore, we can learn from the general image super-resolution task, in which many well-designed network structures have been continuously proposed (IPT \cite{chen2021pre} and SwinIR \cite{liang2021swinir}), and design an effective deep network that is more suitable for FSR task. In addition to the effectiveness, an efficient network is also needed in practice, where the large model (with a mass of parameters and high computation costs) is very difficult to be deployed in real-world applications. Hence, developing models with lighter structure and lower computational taxing is still a major challenge.

\textbf{Exploitation of Facial Prior.}
As a domain-specific super-resolution technique, FSR can be used to recover the facial details which are lost in the observed LR face images. The key to the success of FSR is to effectively exploit the prior knowledge of human faces, from 1D vector (identity and attributes), to 2D images (facial landmarks, facial heatmaps and parsing maps), and to 3D models. Therefore, discovering new prior knowledge of human face, how to model or represent these prior knowledge, and how to integrate this information organically into the end-to-end training framework are worthy of further discussion. In addition to these explicit prior knowledge, how to model and utilize the implicit prior that is learned from the data (such as the GAN prior \cite{StyleGAN,stylegan2}), may be another direction.

\textbf{Metrics and Loss Functions.}
As we know, the pixel-wise $\mathcal{L}_{1}$ loss or $\mathcal{L}_{2}$ loss tend to produce the super-resolution results with high PSNR and SSIM values, while perceptual loss and adversarial loss are in favor of letting the model produce some visually pleasant results, \emph{i.e.}, good performance in terms of LPIPS and FID. Therefore, the assessment metric plays an important role in guiding the model optimization and affecting the final results. If we want to obtain a trustable result (in criminal investigation application), PSNR and SSIM may be better metrics. In contrast, if we just want some visually pleasant results, employing LPIPS and FID metrics may be a good choice.
As a result, there is no universal assessment metric that can make the best of both worlds. Therefore, assessment metrics for FSR need more exploration in the future.

\textbf{Discriminate FSR.}
In most situations, our goal is not only to reconstruct a visually pleasing HR face image. Actually, we hope that the super-resolved results can improve the face recognition task by human or computer. Therefore, it would be beneficial to recover a discriminated HR face image (for human) or discriminated feature (for computers) from an LR face image. To enhance the discriminant of super-resolved face images, we can use the weakly-supervised information (paired positive or negative samples) of the training sample to force the model to be able to reconstruct a discriminative face image.

\textbf{Real-world FSR.}
The degradation process in the real world is too complex to be simulated, which results in a large gap between the synthesized LR and HR pairs and real-world data. When applying models trained by synthesized pairs to real-world LR face images, their performance drops dramatically. Given the HR training face images and the unpaired real-world LR face images, some methods \cite{LRGAN, goswami2020robust, aakerberg2021real} have been proposed to learn the real image degradation to create the sample pairs of synthesis LR face images and HR face images. These methods achieve better performance than previous approaches trained with the data produced by bicubic degradation. These methods actually have a potential assumption that all real-world LR face images share the same degradation, \emph{i.e.}, captured from the same camera. However, the obtained real-world LR face images are very different, and their degradation processes are different. Therefore, designing a more robust real-world FSR method is one of the problem has to be settled urgently.

\textbf{Multi-modal FSR.}
Due to the rapid development of sensing technology, multiple sensors in the same system, such as autonomous driving and robots, are becoming more and more common. The utilization of multi-modal information (including audio, depth, near infrared) will be increasingly promoted. Evidently, different modalities provide different clues. In this field, researchers always explore image-related information, such as attribute, identity, and others. Nevertheless, the emergence of audio-guided FSR~\cite{FSA} and hyperspectral FSR \cite{jiang2021hfsr} inspire us to take advantage of information belonging to different modalities. This trend will undoubtedly continue and diffuse into every category in this field. The introduction of multi-modal information will also spur the development of FSR.

\bibliographystyle{ACM-Reference-Format}
\bibliography{sample-short}

\end{document}